\documentclass[10pt,conference]{IEEEtran}

\usepackage[]{graphicx}    
\newcommand{\ignore}[1]{}  

\usepackage[absolute]{textpos}

\usepackage{hyperref}
\hypersetup{pdfborder={0 0 0},colorlinks=true,urlcolor=blue,linkcolor=blue,citecolor=blue,bookmarks=true}
\hypersetup{pdftitle={Reducing Object Detection Uncertainty from RGB and Thermal Data for UAV Outdoor Surveillance}}
\hypersetup{pdfauthor={Juan Sandino, Peter A. Caccetta, Conrad Sanderson, Frederic Maire, Felipe Gonzalez}}

\usepackage{array}
\usepackage{amsmath}
\usepackage{amssymb}
\usepackage[utf8]{inputenc}
\usepackage[T1]{fontenc}
\usepackage{newtxtext}   
\usepackage{inconsolata} 
\usepackage{microtype}
\DisableLigatures[f]{encoding = *, family = * }

\usepackage[capitalise,noabbrev]{cleveref}  
\usepackage{algorithm}
\usepackage{algpseudocode}
\usepackage{setspace}
\usepackage{float}
\usepackage{tikz}
\usepackage[backgroundcolor=green!20]{todonotes}
\usepackage{soul}
\usepackage{enumerate}

\newcommand{\UAV}{u}
\newcommand{\Victim}{v}
\newcommand{\Step}{k}
\newcommand{\States}{S}
\newcommand{\Actions}{A}
\newcommand{\Observations}{O}
\newcommand{\RewardFun}{R}
\newcommand{\TransitionFun}{T}
\newcommand{\Policy}{\pi}
\newcommand{\DiscountFactor}{\gamma}
\newcommand{\Time}{t}

\newcommand{\ObservationFun}{{\mathcal{Z}}}
\newcommand{\Belief}{b}

\newcommand{\History}{H}

\newcommand{\ie}{{i.e.}}
\newcommand{\eg}{{e.g.}}

\begin{document}
\title{\LARGE\bf Reducing Object Detection Uncertainty from RGB and Thermal Data for UAV Outdoor Surveillance}

\author{%
  Juan Sandino\textsuperscript{{\tiny~}$\diamond\ddagger$},
  Peter A. Caccetta\textsuperscript{{\tiny~}$\ddagger$},
  Conrad Sanderson\textsuperscript{{\tiny~}$\dagger\ddagger$},
  Frederic Maire\textsuperscript{{\tiny~}$\diamond$},
  Felipe Gonzalez\textsuperscript{{\tiny~}$\diamond$}\\
  ~\\
  \textsuperscript{$\diamond$}{\tiny~}Queensland University of Technology, Australia\\
  \textsuperscript{$\dagger$}{\tiny~}Griffith University, Australia\\
  \textsuperscript{$\ddagger$}{\tiny~}Data61~/~CSIRO, Australia
}

\maketitle
\thispagestyle{empty}
\pagestyle{empty}

\begin{abstract}
  Recent advances in Unmanned Aerial Vehicles (UAVs) have resulted in their quick adoption for wide a range of civilian applications, including precision agriculture, biosecurity, disaster monitoring and surveillance.
  UAVs offer low-cost platforms with flexible hardware configurations, as well as an increasing number of autonomous capabilities, including take-off, landing, object tracking and obstacle avoidance.
  However, little attention has been paid to how UAVs deal with object detection uncertainties caused by false readings from vision-based detectors, data noise, vibrations, and occlusion.
  In most situations, the relevance and understanding of these detections are delegated to human operators, as many UAVs have limited cognition power to interact autonomously with the environment.
  This paper presents a framework for autonomous navigation under uncertainty in outdoor scenarios for small UAVs using a probabilistic-based motion planner.
  The framework is evaluated with real flight tests using a sub 2 kg quadrotor UAV and illustrated in victim finding Search and Rescue (SAR) case study in a forest/bushland.
  The navigation problem is modelled using a Partially Observable Markov Decision Process (POMDP), and solved in real time onboard the small UAV using Augmented Belief Trees (ABT) and the TAPIR toolkit.
  Results from experiments using colour and thermal imagery show that the proposed motion planner provides accurate victim localisation coordinates, as the UAV has the flexibility to interact with the environment and obtain clearer visualisations of any potential victims compared to the baseline motion planner.
  Incorporating this system allows optimised UAV surveillance operations by diminishing false positive readings from vision-based object detectors.
\end{abstract}

\begin{textblock}{13.4}(1.3,14.9)
\hrule
\vspace{1ex}
\noindent
\scriptsize
\textbf{{$^\ast$}~Published in:} IEEE Aerospace Conference, 2022. DOI:~\href{https://doi.org/10.1109/AERO53065.2022.9843611}{\tt 10.1109/AERO53065.2022.9843611}
\end{textblock}

\section{Introduction}
Recent advances in autonomous navigation of Unmanned Aerial Vehicles (UAVs)---also known as drones---have resulted in their gradual adoption in a set of civilian and time-critical applications such as surveillance, disaster monitoring, and Search and Rescue (SAR)~\cite{Hanson2017,Lee2016a,Motlagh2017,Pajares2015}.
UAVs offer unique benefits such as compact sizes and low cost to scout outdoor and indoor environments, real-time telemetry and camera streaming to monitor challenging and otherwise inaccessible environments, extensive payload adaptability, and extensive possibilities to augment navigation capabilities through software~\cite{Erdelj2016,JimenezLopez2019,Pensieri2020,Sandino2020}.

\begin{figure}[!b]
  \centering
  \includegraphics[width=\columnwidth]{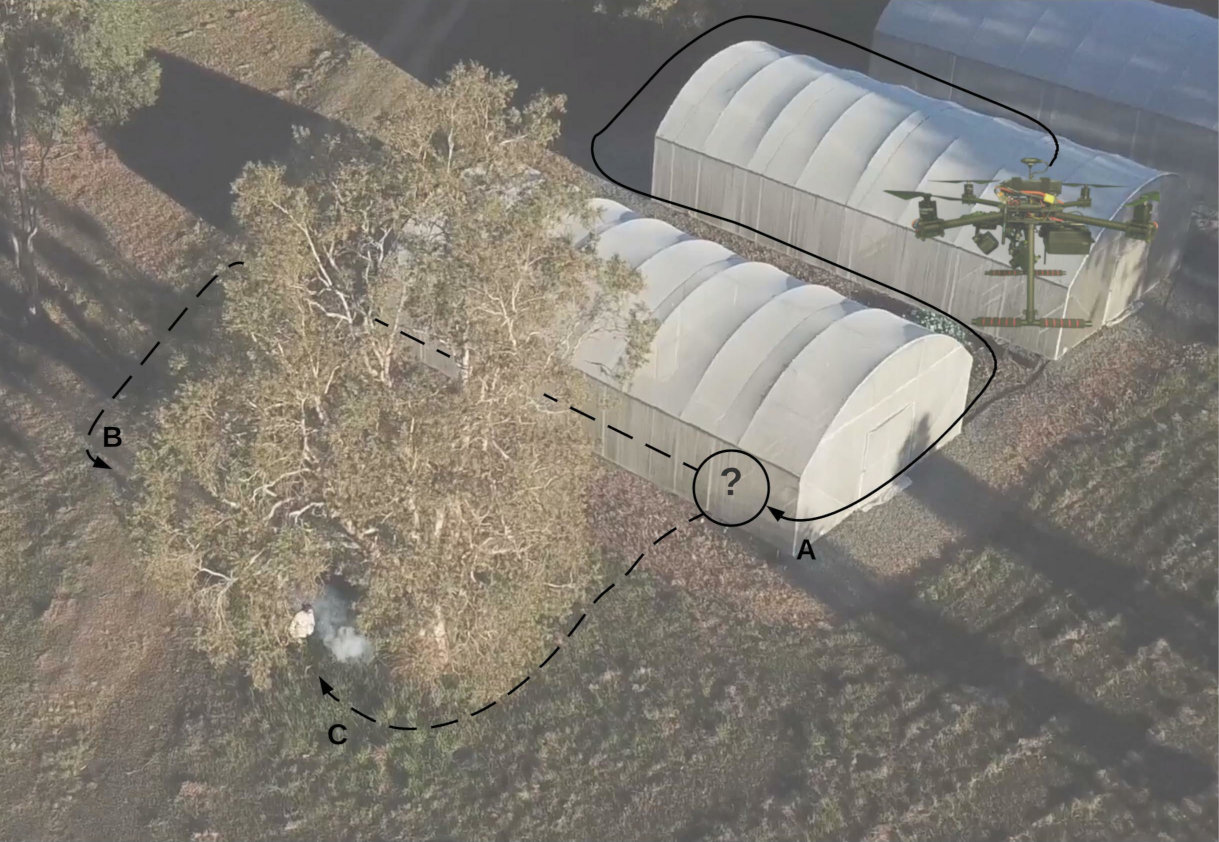}
  \vspace{-4ex}
  \caption{Unmanned aerial vehicle (UAV) navigating in environments under uncertainty and partial observability.
    A small UAV with autonomous decision-making should be able to plan sequential sets of actions for optimal navigation trajectories, despite limitations from imperfect sensor data.}
  \label{fig:context}
\end{figure}

One critical challenge in deploying UAVs and robots in general into real-world and time-critical applications is the ever-presence of uncertainty.
Factors that cause uncertainty are diverse, and they can be classified as external or internal.
External factors come from sources beyond the scope of the UAV, such as poor weather and illumination conditions, strong gusts, unknown situational-awareness of surveyed environments, and partial observability.
Internal factors include sub-optimal camera calibration settings, low image resolution, noisy camera frames during streaming, or imperfect detection outputs from computer vision detectors.
As shown in Fig.~\ref{fig:context}, uncertainty sources that are poorly managed can compromise the behaviour of UAVs and the flight mission itself~\cite{Macdonald2018}.
Thus, it is essential to incorporate cognitive capabilities in UAVs to broaden their use in more real-world scenarios~\cite{Hassanalian2018}.

The elevated number of stranded people and human loss is a problem that is far from solved~\cite{ANSR2020}.
In Australia alone, an average of 38,000 people per year are reported missing and around 2\% of them (or 720 persons) are never located~\cite{Bricknell2017}.
In the event of an emergency---where time management plays a critical factor in the success of the rescue operation---the goal to identify and locate as many victims as quick as possible.
Thus, UAV technology for autonomous navigation and victim detection in challenging environments could assist first-responders in locating as many victims as soon as possible.

Research works on applied decision-making theory in UAVs is extensive and indicates that using Partially Observable Markov Decision Processes (POMDPs) onboard UAVs can increase their cognitive capabilities for autonomous navigation and object detection under uncertainty~\cite{Chen2016b,Ilhan2016,CarvalhoChanel2018}.
UAV frameworks for object detection and tracking have been tested in cluttered indoor environments and in the absence of Global Navigation Satellite System (GNSS) coverage \cite{Sandino2020,Sandino2020a,Vanegas2016a,Vanegas2016d}.
POMDPs have also been applied to solve multi-objective problems in UAVs, addressing tasks such as path planning, multiple object detection and tracking, and collision prevention~\cite{Ragi2013,Ragi2015}.

In time-critical applications such as SAR, real-time camera streaming is critical to comprehend the context of the environment~\cite{Mayer2019}.
However, drone pilots have a strong reliance on their communication systems to control most UAVs.
If communication systems fail, the usability of the UAV could be seriously compromised~\cite{Valavanis2015b}.
Many approaches of POMDPs applied in UAVs for humanitarian relief operations have been tested in simulation \cite{Bravo2018,Waharte2010} and very few systems have been evaluated with real flight test using trivial targets~\cite{Gupta2017b}.

Research efforts on onboard decision-making under object detection uncertainty from Convolutional Neural Network (CNN) models are scarce.
Research conducted in \cite{Sandino2020,Sandino2020a} described a framework and POMDP problem formulation for a SAR application in GNSS-denied environments with a sub 2~kg UAV.
However, the framework was only tested in cluttered indoor scenarios.

This paper describes a modular UAV framework for autonomous onboard navigation in outdoor environments under uncertainty.
The framework design aims to reduce levels of object detection uncertainty using a POMDP-based motion planner, which allows the UAV to interact with the environment to obtain better visual representations of detected objects.
CNN-based computer vision inference and motion planning can be executed in resource-constrained hardware onboard small UAVs.
The framework is tested with real flight tests with a simulated SAR mission, which consisted of finding an adult mannequin in an open area and close to a tree.
Three flight modes are proposed to evaluate the feasibility of the framework for real-world SAR operations.

This paper extends the work in \cite{Sandino2021,Sandino2020,Sandino2020a} with the following contributions:
(1) an extension of their evaluated UAV framework---originally designed for navigation in GNSS-denied environments--- for outdoor missions with GNSS signal coverage, and the design of a novel flight mode;
(2) an additional validation of preliminary results of their proposed UAV framework with comprehensive real flight tests;
and (3) a scalability approach of the framework by adapting a thermal camera and a custom object detector to locate victims using their heat signatures.

The rest of the paper is structured as follows.
\Cref{sec:framework-design} details the UAV framework design for autonomous object detection in uncertain outdoor environments.
\Cref{sec:planner} summarises the implemented probabilistic-based motion planner using a POMDP.
The design of conducted experiments using real flight tests is presented in \Cref{sec:experiments}.
Obtained results and discussion of performance indicators are provided in \Cref{sec:results}.
Conclusions and future avenues for research are discussed in \Cref{sec:conclusions}.

\section{Framework Design}\label{sec:framework-design}
\begin{figure*}[!ht]
  \centering
  \includegraphics[width=1.7\columnwidth]{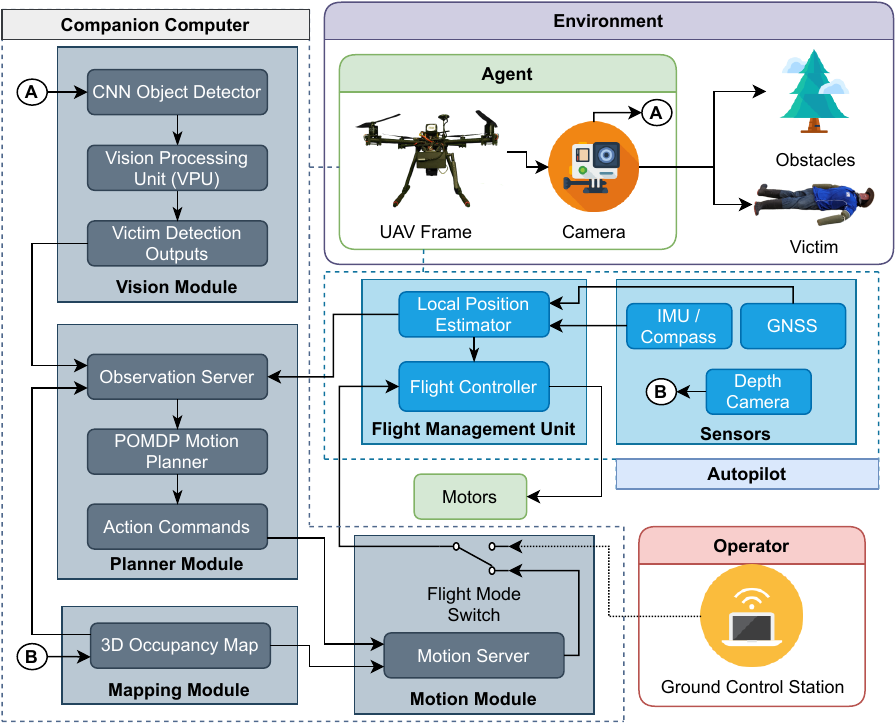}
  \caption{Modular system architecture for autonomous navigation onboard UAVs in uncertain outdoor environments.
    The framework portrays the physical environment (or world) composed of the UAV frame, attached payloads, world obstacles, and the victim.
    A companion computer is attached to the UAV to execute software algorithms in dedicated modules for computer vision, mapping, real-time path planning, and a motion server that interfaces the companion computer with the UAV autopilot.
  }
  \label{fig:sys_arch}
\end{figure*}

The framework follows a modular system architecture for autonomous navigation onboard small UAVs as illustrated in Fig.~\ref{fig:sys_arch}.
This design extends an existing UAV framework for autonomous navigation in cluttered environments under object detection uncertainty, tested in simulation and with real flight tests in a sub 2~kg quadcopter~\cite{Sandino2020a}.

Fig.~\ref{fig:sys_arch} illustrates the physical environment (or world) composed by the UAV frame and any attached payloads (\ie, RGB or thermal cameras), the victim and obstacles.
Acquired camera frames represent the visual interface (also called observations) of the surveyed environment by the UAV.
The UAV also contains the \emph{autopilot}, which translates high-level action commands into low-level signals that control the UAV motors.
The last hardware component of the UAV frame is a companion computer, which is allocated to execute software algorithms in dedicated modules for computer vision, mapping, and real-time path planning.
Action commands from the planner are managed by the motion module, which interfaces with the flight controller of the autopilot.

The following subsections discuss each of the proposed framework components.
The UAV framework used in this work is not limited to the hardware and software discussed below.
Other UAV frame designs, payloads, autopilots, vision-based object detectors, planners, and software toolkits can also be implemented.

\subsection{UAV Airframe and Payloads}\label{sec:uav_frame}
\begin{figure}[!tb]
  \centering
  \resizebox{\columnwidth}{!}
  {
  \tikzset{every picture/.style={line width=0.75pt}} 
  \begin{tikzpicture}[x=0.75pt,y=0.75pt,yscale=-1,xscale=1]

  \draw (195,134.89) node  {\includegraphics[width=292.5pt,height=187.34pt]{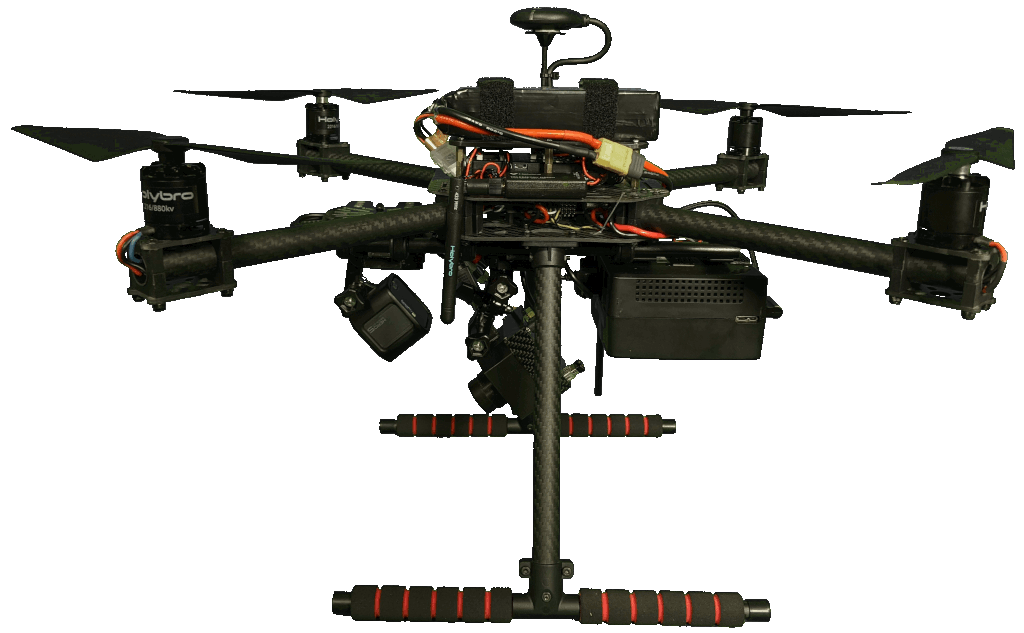}};
  \draw [color={rgb, 255:red, 208; green, 2; blue, 27 }  ,draw opacity=1 ]   (20,80) -- (67,80) ;
  \draw [shift={(70,80)}, rotate = 180] [fill={rgb, 255:red, 208; green, 2; blue, 27 }  ,fill opacity=1 ][line width=0.08]  [draw opacity=0] (10.72,-5.15) -- (0,0) -- (10.72,5.15) -- (7.12,0) -- cycle    ;
  \draw [color={rgb, 255:red, 208; green, 2; blue, 27 }  ,draw opacity=1 ]   (280,30) -- (212.6,68.51) ;
  \draw [shift={(210,70)}, rotate = 330.26] [fill={rgb, 255:red, 208; green, 2; blue, 27 }  ,fill opacity=1 ][line width=0.08]  [draw opacity=0] (10.72,-5.15) -- (0,0) -- (10.72,5.15) -- (7.12,0) -- cycle    ;
  \draw [color={rgb, 255:red, 208; green, 2; blue, 27 }  ,draw opacity=1 ]   (170,20) -- (197,20) ;
  \draw [shift={(200,20)}, rotate = 180] [fill={rgb, 255:red, 208; green, 2; blue, 27 }  ,fill opacity=1 ][line width=0.08]  [draw opacity=0] (10.72,-5.15) -- (0,0) -- (10.72,5.15) -- (7.12,0) -- cycle    ;
  \draw [color={rgb, 255:red, 208; green, 2; blue, 27 }  ,draw opacity=1 ]   (100,120) -- (167.12,100.82) ;
  \draw [shift={(170,100)}, rotate = 524.05] [fill={rgb, 255:red, 208; green, 2; blue, 27 }  ,fill opacity=1 ][line width=0.08]  [draw opacity=0] (10.72,-5.15) -- (0,0) -- (10.72,5.15) -- (7.12,0) -- cycle    ;
  \draw [color={rgb, 255:red, 208; green, 2; blue, 27 }  ,draw opacity=1 ]   (350,140) -- (293,140) ;
  \draw [shift={(290,140)}, rotate = 360] [fill={rgb, 255:red, 208; green, 2; blue, 27 }  ,fill opacity=1 ][line width=0.08]  [draw opacity=0] (10.72,-5.15) -- (0,0) -- (10.72,5.15) -- (7.12,0) -- cycle    ;
  \draw [color={rgb, 255:red, 208; green, 2; blue, 27 }  ,draw opacity=1 ]   (80,180) -- (177.06,160.59) ;
  \draw [shift={(180,160)}, rotate = 528.69] [fill={rgb, 255:red, 208; green, 2; blue, 27 }  ,fill opacity=1 ][line width=0.08]  [draw opacity=0] (10.72,-5.15) -- (0,0) -- (10.72,5.15) -- (7.12,0) -- cycle    ;
  \draw [color={rgb, 255:red, 208; green, 2; blue, 27 }  ,draw opacity=1 ]   (80,180) -- (127.66,141.87) ;
  \draw [shift={(130,140)}, rotate = 501.34] [fill={rgb, 255:red, 208; green, 2; blue, 27 }  ,fill opacity=1 ][line width=0.08]  [draw opacity=0] (10.72,-5.15) -- (0,0) -- (10.72,5.15) -- (7.12,0) -- cycle    ;

  \draw (13.5,78) node  [font=\large,color={rgb, 255:red, 208; green, 2; blue, 27 }  ,opacity=1 ] [align=left] {\textbf{1}};
  \draw (286.5,28) node  [font=\large,color={rgb, 255:red, 208; green, 2; blue, 27 }  ,opacity=1 ] [align=left] {\textbf{3}};
  \draw (165.5,18) node  [font=\large,color={rgb, 255:red, 208; green, 2; blue, 27 }  ,opacity=1 ] [align=left] {\textbf{2}};
  \draw (93.5,122) node  [font=\large,color={rgb, 255:red, 208; green, 2; blue, 27 }  ,opacity=1 ] [align=left] {\textbf{4}};
  \draw (356.5,138) node  [font=\large,color={rgb, 255:red, 208; green, 2; blue, 27 }  ,opacity=1 ] [align=left] {\textbf{5}};
  \draw (74.5,178) node  [font=\large,color={rgb, 255:red, 208; green, 2; blue, 27 }  ,opacity=1 ] [align=left] {\textbf{6}};
  \end{tikzpicture}
  }
  \caption{Framework implementation in a sub 2 kg quadrotor UAV.
    Primary components include:
    (1) carbon fibre Holybro X500;
    (2) Pixhawk~4 GNSS receiver;
    (3) Pixhawk~4 autopilot;
    (4) 433 MHz telemetry radio;
    (5) Intel UP\textsuperscript{2} companion computer;
    (6) payload.}
  \label{fig:uav}
\end{figure}
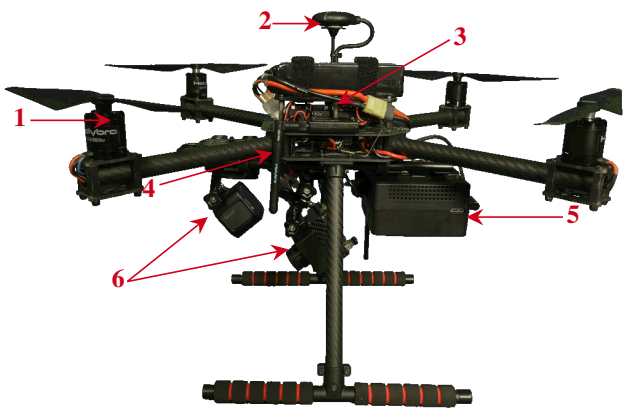

The UAV airframe which offered the best combination between payload adaptability, size, and endurance for this research is a Holybro X500 quadrotor kit (Holybro, China).
As shown in Fig.~\ref{fig:uav}, key components utilised from the kit include a Pixhawk~4 autopilot, Pixhawk~4 GNSS receiver, 2216~KV880 brushless motors, 22.86~cm plastic propellers, and a 433~MHz Telemetry Radio.
With dimensions of 41~cm \(\times\) 41~cm \(\times\) 30.0~cm, the UAV carries a four cell 5000~mAh LiPo battery, for an approximate flight autonomy of 12 min.

The companion computer is an Intel UP\textsuperscript{2},
chosen for its price tag, number of peripherals and CPU architecture.
Key specifications include a 64-bit quad-core CPU at 1.1 GHz, 64 GB eMMC SSD, 8 GB RAM, four FL110 USB 3.0 connectors, two High-Speed UART controllers, and one mPCIe connector.

The proposed framework was tested using two Red, Green, Blue (RGB) cameras, namely an Arducam B019701 and a GoPro Hero 9.
Thermal imagery is sourced from a FLIR Tau 2 connected to a ThermalCapture device for real-time frame streaming.
The cameras, which can be interchangeably used in the proposed framework, are mounted onto an anti-vibration bracket, pointing to the ground and in parallel to Earth's nadir, as seen in Fig.~\ref{fig:gph9}.
Core properties for the cameras can be found in Tab.~\ref{tab:survey_params_rgb} in the Appendix.

\begin{figure}[!tb]
  \centering
  \includegraphics[width=\columnwidth]{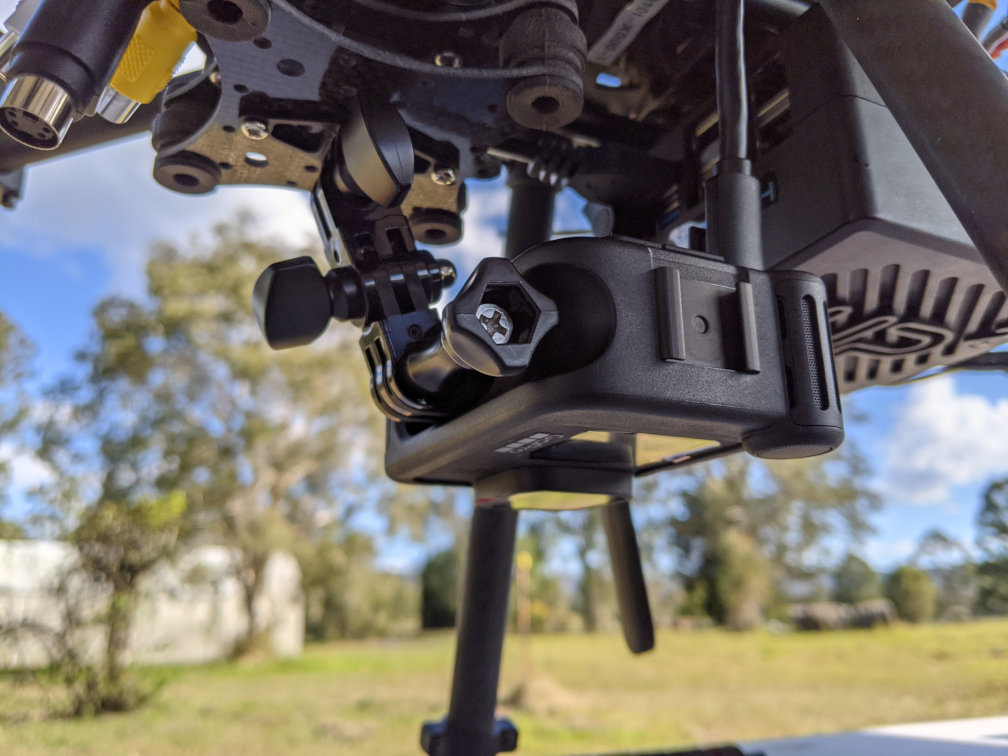}
  \caption{GoPro Hero 9 mounted onto an anti-vibration bracket, pointing to the ground and in parallel to Earth's nadir.}
  \label{fig:gph9}
\end{figure}

\subsection{Vision Module}
This module consists of a deep learning object detector processing raw frames from the GoPro Hero 9 camera.
Taking into account the performance limitations of running deep learning models in resource-constrained hardware, a Vision Processing Unit (VPU) is installed in the companion computer.
Convolutional operations that normally run onboard a CPU or GPU are allocated to the VPU for inference of CNN models in resource-constrained hardware.
In this implementation, the selected VPU is an Intel Movidius Myriad X, which is connected to the companion computer via the mPCIe slot.
The~detection module is programmed in Python and uses the OpenVINO library to optimise code instructions to load CNN models into the VPU.

The deep learning model architecture used to detect victims is an off-the-shelf Google MobileNet Single-Shot Detector (SSD)~\cite{Chuanqi2020}.
This model is deployed in Caffe~\cite{Jia2014} and tuned with pre-trained weights from the PASCAL VOC2012 dataset~\cite{Everingham2015}, scoring a mean average precision of 72.7\%.
The~dataset covers up to 21 class objects (including persons).
However, only positive detections for the class {person} are evaluated.
Acquired camera frames are fitted into the input layer of the neural network (i.e.,~MobileNet SSD model) by downsizing the frames to 300 \( \times \) 300 pixels.

\subsection{Mapping Module}
The Mapping module manages 3D occupancy maps, which are constituted by volumetric occupancy grids and displays the presence and localisation of objects in the surveyed environment.
In this implementation, the 3D Occupancy Map are requested by the Motion and Planner modules to evaluate the presence of obstacles at selected position coordinates in the world coordinate frame.
The maps are created through the use of the Octomap library \cite{Hornung2013}.

\subsection{Planner Module}
The planner module computes the motion policy of the UAV and contains three primary components:
(1) an observation server, which handles raw observations from the vision module (\ie, detected victims, confidence and victim coordinates), local position estimations of the UAV, and the state of the 3D occupancy map;
(2) the POMDP motion planner, which calls the observation server every time the planner requires new observations;
and (3) action commands computed by the motion policy of the planner and are read by the motion server.
Complete details of the POMDP planner design can be found in \Cref{sec:planner}.




\subsection{Communication Interface}
The UAV framework runs on open-source software tools.
The companion computer runs under Linux Ubuntu 18.04 LTS O.S. and the Robot Operating System (ROS) melodic.
ROS is a middleware to communicate between the nodes of each module (following the architecture design from Fig.~\ref{fig:sys_arch}).
The Pixhawk~4 autopilot is powered by PX4, which communicates with the companion computer through MAVROS, a ROS implementation of the MAVLink protocol, which is industry standard for UAV communication and control~\cite{Koubaa2019}.
The POMDP solver implementation, which is described in \Cref{sec:planner}, also contains a ROS implementation to maximise the use of visualisation, telemetry and recording tools from ROS.

\section{Planner Design}\label{sec:planner}
This approach formulates the decision-making problem as a POMDP.
The planner transmits UAV position commands to the motion planner derived from environment observations.
The discussion presented in this section is adapted from \cite{Sandino2021} and only essential parts are shown in this paper for completeness.

With a taken action \(a \in \Actions\), the UAV receives an observation \(o \in \Observations\) encoded by the observation function \( {\ObservationFun}(s^{\prime},a, o) = {\mathbb{P}}\left(o \mid s^{\prime}, a\right) \).
Every decision chain is then quantified with an estimated reward \(r\), calculated using the reward function \(\RewardFun(a, s)\).
A POMDP uses a probability distribution over the system states to model uncertainty of its observed states.
This modelling is called the belief \( \Belief (\History) = {\mathbb{P}} [s^{1} \mid \History ],~\cdots~, {\mathbb{P}} [ s^{n} \mid \History]\), where \( \History \) is the history of actions, observations and rewards the UAV has accumulated until a time step \(\Time\), or \(\History= a_{0} , o_{1}, r_{1} ,~\cdots~, a_{\Time - 1} , o_{\Time}, r_{\Time}\).

The motion policy \(\Policy \) of the UAV is represented by mapping belief states to actions \(\Policy: \Belief \rightarrow \Actions \).
A POMDP is solved after finding the optimal policy \(\Policy^{*}\), calculated as follows:
\begin{equation}
  \Policy^{*}:=\underset{\pi}{\arg \max }\left({\mathbb{E}}\left[\sum_{\Time=0}^{\infty} \DiscountFactor^{\Time} \RewardFun\left(\States_{\Time}, \Policy\left(b_{\Time}\right)\right) \right]\right)
\end{equation}

\noindent where \( \DiscountFactor \in [0,1] \) is the discount factor and defines the relative importance of immediate rewards compared to future rewards.
A given POMDP solver starts planning from an initial belief \(\Belief_{0}\), which is usually generated using the initial conditions (and assumptions) of the flight mission.

\subsection{Assumptions}
In this implementation, the formulated problem for exploration and object detection (\ie, victims) using multi-rotor UAVs in outdoor environments assumes:

\begin{small}
\begin{enumerate}[{$\bullet$}]
\itemsep=1ex
  \item An initial 3D occupancy map of the environment is pre-loaded to the planner before the UAV takes off.
  \item Observations come from processed camera frames (by the Vision module), the 3D occupancy map (by the Mapping Module),  and the estimated local UAV position (by the Autopilot).
  \item Only a single, and static victim can be detected at the same time.
        If more victims appear on processed camera frames, the planner will only read data from the victim with the highest detection confidence values.
  \item The motion planner starts once the UAV reaches known position setpoint (\ie, at one of the corners of the surveyed area).
  \item The planner stops computing a motion policy once:
        (1)~the~UAV detects a victim whose detection confidence surpasses a set threshold;
        (2)~the~UAV explores the whole search area extent without finding any victims;
        or (3)~the UAV exceeds the maximum flight time on air (because of low levels of battery power).
\end{enumerate}
\end{small}

\subsection{Actions}
UAV actions are defined by seven position commands, namely \textit{forward, backward, left, right, up, down,} and \textit{hover}.
UAV actions that are not included in the action space but are managed by the autopilot instead include \textit{arm, disarm, take-off, return to launch and land}.
Each action updates the position set point of the UAV in the world coordinate frame by calculating and applying a change of position \(\delta\).

The magnitude for \( \delta_{x} \)  and \( \delta_{y} \) depends on the  estimated overlap value between camera frame observations:
%
\begin{equation}
  \delta = l_{\text{FOV}}(1 - \lambda ) \label{eq:delta-pos}
\end{equation}

\noindent where \(l_{\text{FOV}}\) is the length of the projected camera's Field of View (FOV),
and \(\lambda \in [0, 1) \) is the desired overlap value.

\subsection{States}
A system state \( s \in \States \) is defined as:
\begin{equation}
  s = (p_{\UAV}, f_{\text{roi}}, f_{\text{dct}}, p_{\Victim}, c_{\Victim})
\end{equation}

\noindent where
\(p_{\UAV}\) is the position of the UAV in the world coordinate frame,
\(f_{\text{crash}}\) is a flag raised when the UAV crashes with an obstacle,
\(f_{\text{roi}}\) is a flag indicating whether the UAV is flying beyond the flying limits,
\(f_{\text{dct}}\) is the flag raised if a potential victim is detected by the UAV.
If \(f_{\text{dct}} = \text{True}\), the position of the victim in the world coordinate frame
is given in \(p_{\Victim}\),
with detection confidence \(c_{\Victim}\in [0, 1]\).
The system  reaches a terminal state whenever \(c_{\Victim} \geq \zeta\), where \(\zeta\) is the confidence threshold.

\subsection{Transition Function}
The motion dynamics of a multi-rotor UAV defines the transition from current to new states:
\begin{equation}
  p_{\UAV}(\Step + 1) = p_{\UAV}(\Step) + \Delta p_{\UAV}(\Step)\label{eq:motion-model}
\end{equation}

\noindent where \(p_{\UAV}(\Step)\) is the position of the UAV at time step \(k\),
and \(\Delta p_{\UAV}(\Step)\) is the position change of the UAV between time steps.
This formulation does not contain any actions for heading changes. However, Eqn.~\ref{eq:motion-model} can be expanded if required by adding the rotation matrix in multi-copters~\cite{Chovancova2014}.
An illustration of a problem formulation including the rotation matrix can be found in~\cite{Sandino2020}.

\subsection{Reward Function}\label{sec:rewards}
The expected reward \( r \) after taking an action \(a \in \Actions\) from state \(s \in \States\) is calculated using the reward function \(\RewardFun(a,s)\) defined in \Cref{alg:reward_fx,tab:reward_vars}.
This function critically influences the UAV behaviour during flight missions, and its definition allows multi-objective task definition.
A complete discussion on the design considerations of the reward function can be found in \cite{Sandino2021}.

\begin{small}
\begin{algorithm}[!tb]
  \caption{Reward function \(\RewardFun\) for exploration and object detection in outdoor environments.}\label{alg:reward_fx}
  \begin{algorithmic}[1]
    \State $r \gets 0$
    \If{$f_{\text{crash}}$}
    \State $r \gets r_{\text{crash}}$\Comment{UAV crashing cost}
    \ElsIf{$f_{\text{roi}}$}
    \State $r \gets r_{\text{out}}$ \Comment{Beyond safety limits cost}
    \ElsIf{$f_{\text{dct}}$}
    \State $r \gets r_{\text{dtc}}$ \Comment{Detected object reward}
    \State $r \gets r \mbox{~+} \left[r_{\text{dtc}} \cdot \left(1 \mbox{~-~} \frac{z_{u} \mbox{-} z_{\text{min}}}{z_{\text{max}} \mbox{-} z_{\text{min}}}\right)\right]$ \Comment{UAV altitude reward}
    \If{$c_{\Victim} \geq \zeta$ \textbf{and} $a = \text{Down}$}
    \State $r \gets r + r_{\text{conf}}$
    \EndIf
    \Else
    \State $r \gets r_{\text{action}}$\Comment{Action cost}
    \State $r \gets r - \left[r_{\text{dtc}} \cdot  \left(1 \mbox{~-~} \frac{z_{u} \mbox{~-~} z_{\text{min}}}{z_{\text{max}} \mbox{~-~} z_{\text{min}}}\right)\right]$ \Comment{UAV altitude cost}
    \State $r \gets r - \left[r_{\text{dtc}} \cdot  \left(1 \mbox{~-~} {0.5}^{{~4 \cdot d_{v}}/{d_{w}}}\right)\right]$\Comment{Horizontal distance cost}
    \State $r \gets r + r_{\text{fov}} \cdot \varepsilon$ \Comment{Footprint overlap cost}
    \EndIf
    \State \Return $r$
  \end{algorithmic}
\end{algorithm}
\end{small}

\begin{table}[!tb]
  \small
  \caption{Applied reward values to the reward function \(\RewardFun\), defined in \Cref{alg:reward_fx}.}
  \label{tab:reward_vars}
  \centering
  \begin{tabular}{|c|r|l|}
  \hline
  \textbf{Variable}     & \textbf{Value} & \textbf{Description}                  \\
  \hline\hline
  \(r_{\text{crash}}\)  & \(-50\)        & Cost of UAV crash                     \\
  \(r_{\text{out}}\)    & \(-25\)        & Cost of UAV breaching safety limits   \\
  \(r_{\text{dtc}}\)    & \(+25\)        & Reward for detecting potential victim \\
  \(r_{\text{conf}}\)   & \(+50\)        & Reward for confirmed victim detection \\
  \(r_{\text{action}}\) & \(-2.5\)       & Cost per action taken                 \\
  \(r_{\text{fov}}\)    & \(-5\)         & Footprint overlapping cost            \\
  \hline
  \end{tabular}
\end{table}

The order of the steps from \Cref{alg:reward_fx} classifies high-level tasks into two components.
The first one is object detection and starts by evaluating any states which will negatively affect the integrity of the UAV, followed by states indicating positive victim detections.
If the UAV detects a potential victim (Step 6), \(\RewardFun\) calculates a linear function (Step 8) which returns increased reward values as the UAV gets closer to the minimum allowed altitude.
A higher reward value is returned if a potential victim is confirmed (Step 9 and 10).

The second component of the algorithm addresses exploration.
In case there are no detections, \(\RewardFun\) applies a set of cost functions to encourage a greedy horizontal exploration of the environment.
An exponential function in Step 14 calculates the Manhattan distance between the UAV and the victim \( d_{v}\) and the maximum exploration distance \( d_{w}\) which are defined as:

\begin{small}
\begin{eqnarray}
  d_{v} & \mbox{=} & \sum\nolimits^{n}_{i=1}{\left| p_{i} - q_{i} \right|}, ~ p_{i} \mbox{=} (x_{\text{\UAV}},y_{\UAV}), ~ q_{i} \mbox{=} (x_{\Victim},y_{\Victim}) \\
  d_{w} & \mbox{=} & \sum\nolimits^{n}_{i=1}{\left| p_{i} - q_{i} \right|}, ~ p_{i} \mbox{=} (x_{\text{max}},y_{\text{max}}), ~ q_{i} \mbox{=} (x_{\text{min}},y_{\text{min}})\label{eq:point-dist}
\end{eqnarray}%
\end{small}%

The overlap \(\varepsilon\) between the camera's current footprint and its correspondent location in the footprint map is defined as:
\begin{equation}
  \varepsilon = \frac{\sum_{i=1}^{n} F_{i}(p_{u})}{n}, ~ \varepsilon \in [0, 1]
  \label{eqn:overlap}
\end{equation}

\noindent where \(F_{i}(p_{u})\) are the pixel values of the projected FOV in the footprint map, and \(n\) is the total number of projected pixels in the footprint.
A maximum overlap value of 1 indicates that such action will place the UAV to a fully previously explored area, and, as indicated in Step 16 of \Cref{alg:reward_fx}, the whole penalty value \(r_{\text{fov}}\) will be added to the reward.
A minimum value of 0 means that a given action will place the UAV in an unexplored area and no penalty will be added to the reward.
Intermediate values of \(\varepsilon\) represent partial overlapping, adding a partial penalty value \(r_{\text{fov}}\) to the reward.

\subsection{Observations}
An observation $o \in \Observations$ is defined as:
\begin{equation}
  o = (o_{p_{\UAV}}, o_{\text{dtc}}, o_{p_{\Victim}}, o_{\zeta}, o_{\text{obs}})
  \label{eqn:observations}
\end{equation}

\noindent where \(o_{p_{\UAV}}\) is the estimated position of the UAV by the autopilot;
\(o_{\text{dtc}}\) is the flag triggered by potential victim detections received by the CNN model;
\(o_{p_{\Victim}}\) and \(o_{\zeta}\) are the local position of the victim and the detection confidence respectively, both of them defined only if there are any positive detections;
and \(o_{\text{obs}}\) is the flag triggered after processing the 3D occupancy map for any obstacles located in front of the UAV.

The detection confidence \(o_{\zeta}\) measures the frequency of positive detections between the last two observation calls:
\begin{equation}
  o_{\zeta} = \dfrac{\sum_{i = 1}^{n} o_{\text{dtc}_{i}}}{n}
  \label{eq:det-conf}
\end{equation}

\noindent where \(n\) is the number of segmented frames between observation calls,
and \(o_{\text{dtc}}\) is the flag indicating a positive detection per processed frame \(i\).

\subsection{Observation Model}\label{sec:obs_model}
This implementation uses Augmented Belief Trees (ABT) \cite{Kurniawati2016}, an online POMDP solver that contains a model that generates \(\TransitionFun\) and \(\ObservationFun\) using a modelled observation \(o\) given an action \(a\) and the next state \(s^{\prime}\).
The variables contained in the generative model are the local position of the UAV \(s^{\prime}_{p_{\UAV}}\), the local position of the victim \(s^{\prime}_{p_{\Victim}}\) and the detection confidence~\(o_{\zeta}\).

\begin{figure}[!b]
  \centering
  \includegraphics[width=\columnwidth]{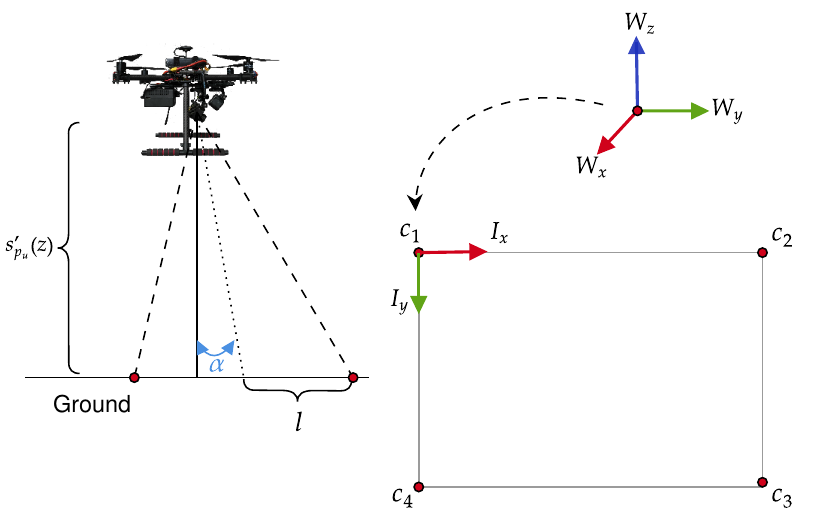}
  \vspace{-5ex}
  \caption{Field of View (FOV) projection and footprint extent of a vision-based sensor. The camera setup on the UAV frame defines \(\alpha \) as the pointing angle from the vertical (or pitch) and determines the coordinates of the footprint corners \(c\).}\label{fig:footprint}
\end{figure}

Potential victim detections and their subsequent positioning estimations are conditioned by the camera pose at the UAV frame and its projected footprint of the environment.
Specifically, if the 2D local position coordinates of the victim \(s^{\prime}_{p_{\Victim}}(x, y)\) are within the projected footprint limits of the camera, the victim is assumed to be detected.
This estimation is done by calculating the sum of angles between a 2D point (\ie{}, \(s^{\prime}_{p_{\Victim}}\)) and each pair of points that constitute the footprint boundaries (the footprint rectangular corners)~\cite{Bourke1997}.
The 2D projected footprint extent \( l \) of a vision-based sensor, illustrated in Fig.~\ref{fig:footprint}, can be calculated using: 
\begin{eqnarray}
  l_{\text{top, bottom}} & = & s^{\prime}_{p_{\UAV}}(z) \cdot \tan\left(\alpha \pm \tan^{-1}\left(\frac{h}{2f}\right)\right)  \label{eq:footprint-1} \\
  l_{\text{left, right}} & = & s^{\prime}_{p_{\UAV}}(z) \cdot \tan\left(\beta  \pm \tan^{-1}\left(\frac{w}{2f}\right)\right)  \label{eq:footprint-2}
\end{eqnarray}

\noindent where \(s^{\prime}_{p_{\UAV}}\) is the UAV altitude,
\(\alpha \) and \(\beta \) are the camera's pointing angles from the vertical \(z\) and horizontal \(x\) axis of the World coordinate frame,
\(w\) is the lens width,
\(h\) is the lens height,
and \(f\) is the focal length.

The footprint corners \(c\) from the camera's local coordinate frame \(I\) are translated to the world's coordinate frame \(W\) using: 
\begin{equation}
  \left[\begin{matrix}
      c'(x) \\
      c'(y)\end{matrix}\right] =
  \left[\begin{matrix}
      s'_{p_{\UAV}}(x) \\
      s'_{p_{\UAV}}(y)\end{matrix}\right] +
  \left[\begin{matrix}
      \cos(\varphi_{\UAV}) & -\sin(\varphi_{\UAV}) \\
      \sin(\varphi_{\UAV}) & \cos(\varphi_{\UAV})  \\
    \end{matrix}\right]
  \left[\begin{matrix}
      c(x) \\
      c(y)\end{matrix}\right] \label{eq:tf}
\end{equation}

\noindent where \(s'_{p_{\UAV}}\) is the next UAV position state,
and \( \varphi_{\UAV} \) is the Euler yaw angle of the UAV.
However, as no actions involve adjusting the heading of the UAV mid-flight, and assuming yaw estimation errors are negligible, Eqn.~\ref{eq:tf} is simplified as: 
\begin{equation}
  \left[\begin{matrix}
      c'(x) \\
      c'(y)\end{matrix}\right] =
  \left[\begin{matrix}
      s'_{p_{\UAV}}(x) + c(x) \\
      s'_{p_{\UAV}}(y) + c(y)\end{matrix}\right] \label{eq:tf-simple}
\end{equation}

The detection confidence \(o_{\zeta}\) that comes as part of the output data from the CNN object detector is modelled using: 

\begin{equation}
  o_{\zeta} = \frac{\left(1 - \zeta_{\text{min}}\right)(d_{\UAV\Victim} - z_{\text{min}} + \zeta_{\text{min}})}{z_{\text{max}} - z_{\text{min}}} \label{eq:zeta}
\end{equation}

\noindent where \(\zeta_{\text{min}}\) is the minimally accepted confidence threshold,
\(z_{\text{max}}\) and \(z_{\text{min}}\) are the maximum and minimum UAV flying altitudes respectively,
and \(d_{\UAV\Victim}\) is the Manhattan distance between the UAV and the victim.

\section{Experiments}\label{sec:experiments}
This research validated the proposed UAV framework with real flight tests on the sub 2 kg quadcopter shown in \Cref{sec:uav_frame}.
The tests were designed under a ground SAR application context, specifically, to locate a lost person last seen around a forest/bushland area.
The subsections below present the location of conducted flights, environment setup, proposed flight modes for data collection and tuned hyperparameters of the online POMDP solver.

\subsection{Location and Environment Setup}
\begin{figure}[!b]
  \begin{minipage}[c]{\columnwidth}
    \centering
    \includegraphics[width=\columnwidth]{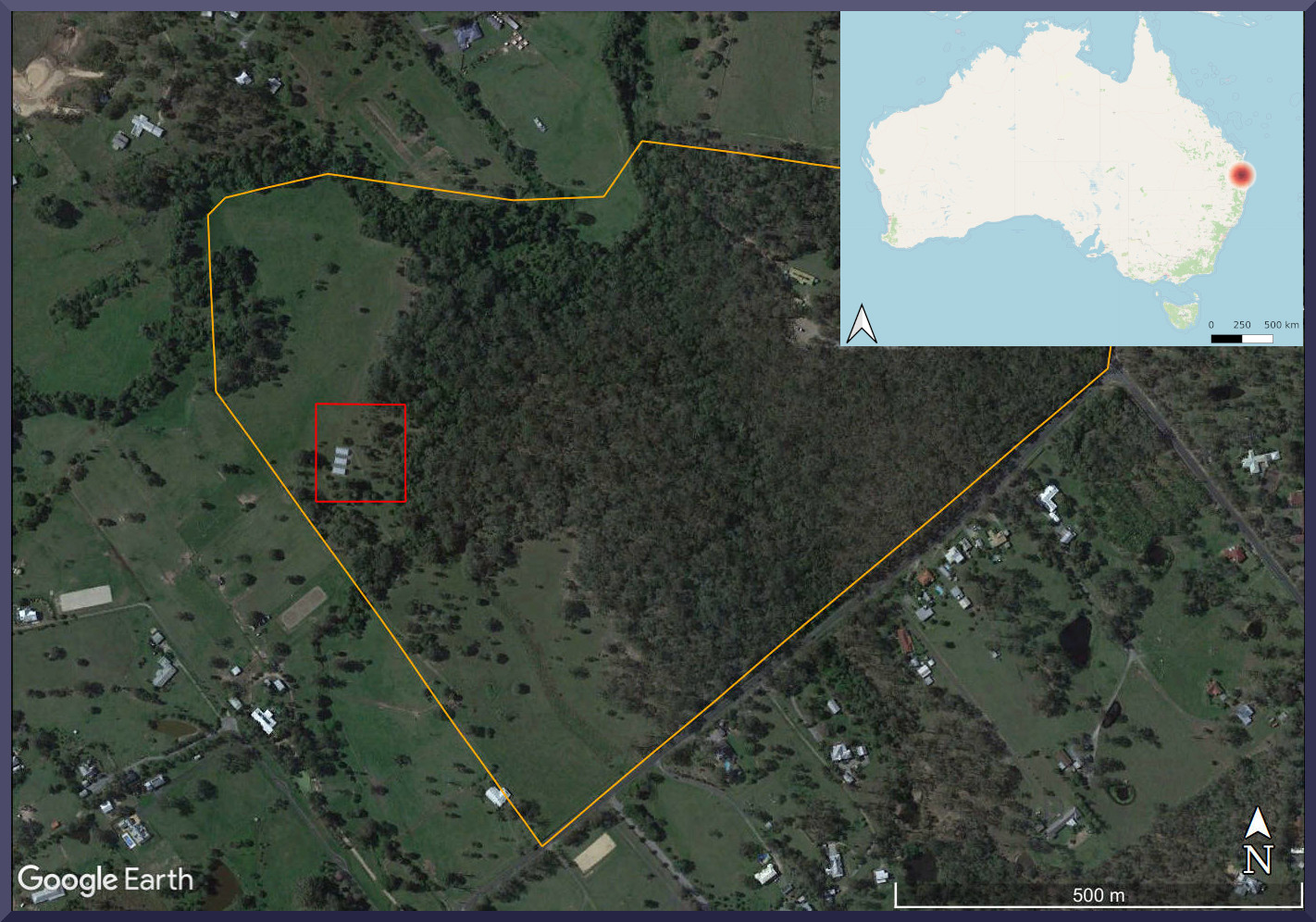}
    \\
    (\textbf{a})
  \end{minipage}
  \\
  \begin{minipage}[c]{\columnwidth}
    \centering
    \includegraphics[width=\columnwidth]{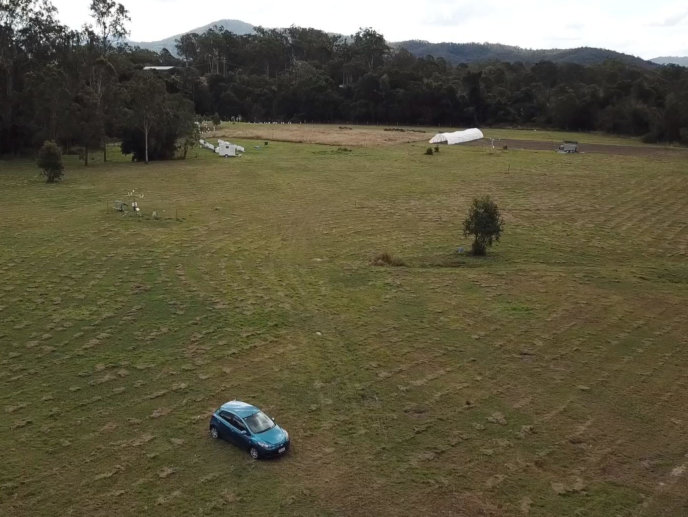}
    \\
    (\textbf{b})
  \end{minipage}
  \centering
  \caption{Location of conducted flight tests at the Samford Ecological Research Facility (SERF), QLD, Australia. (a) SERF and surveyed area boundary extents (orange and red blobs respectively). (b) Aerial footage of surveyed area displaying buffel grassland and obstacles.}
  \label{fig:serf}
\end{figure}

Flight tests were conducted at the Samford Ecological Research Facility (SERF), 148 Camp Mountain Road, Samford QLD 4520, Australia.
As shown in Fig.~\ref{fig:serf}, the 51 hectare property contains protected Dry Sclerophyll forest and grazing zones, where the latter ones were utilised to fly the UAV.

The delimited flying area covers a mostly flat grazing zone featuring buffel grassland, a five-metre tree, and a car purposely placed as an additional obstacle.
Flight tests were conducted between 19 Jul 2021 and 8 Sep 2021, in a rich range of illumination and weather conditions.
Weather conditions included tests under clear and partly cloudy skies, with calm and gusty winds from 6 km/h up to 24 km/h respectively.
The range of recorded temperatures ranged from 14{\textdegree}C to 25{\textdegree}C.

\begin{figure}[!b]
  \begin{minipage}[c]{\columnwidth}
    \centering
    \includegraphics[width=0.85\columnwidth]{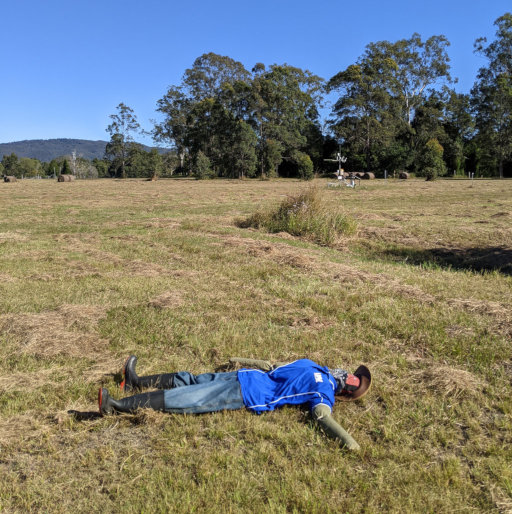}
    \\
    (\textbf{a})\\
    ~\\
  \end{minipage}
  \\
  \begin{minipage}[c]{\columnwidth}
    \centering
    \includegraphics[width=0.85\columnwidth]{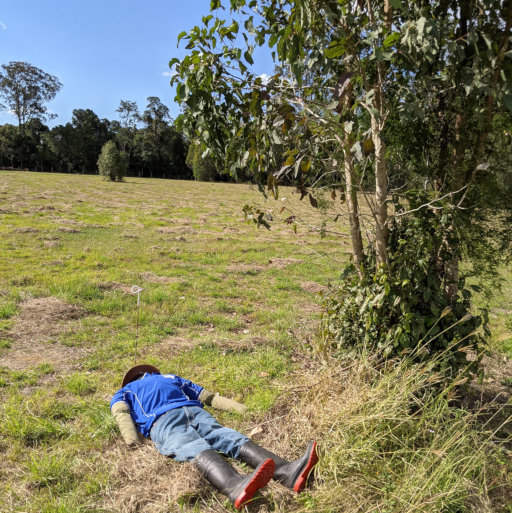}
    \\
    (\textbf{b})
  \end{minipage}
  \centering
  \caption{Adult mannequin placed in the surveyed area as the victim to be found. (a) Trivial victim location (L1) with the mannequin fully exposed in the environment. (b) Complex victim location (L2) with the mannequin partly occluded by a five-metre tree.}
  \label{fig:victim_locations}
\end{figure}

This implementation employed a static adult mannequin posing as the victim to be found for safety reasons.
The mannequin was placed at two predefined locations, as depicted in Fig.~\ref{fig:victim_locations}.
The first location---referred from here as Location~1, or L1---is a trivial setup with the mannequin free of any nearby obstacles and entire visibility from downward-looking cameras.
The second location---referred from here as Location~2, or L2---introduces a complex setup as the mannequin is placed nearby a tree which causes partial occlusion during the flight tests.

\begin{figure*}[!ht]
  \centering
  \includegraphics[width=0.6\textwidth]{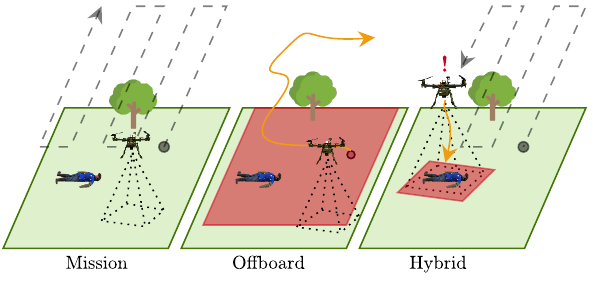}
  \vspace{-5ex}
  \caption{Executed flight modes for exploration and object detection in outdoor environments.
    Mission mode is the baseline motion planner and lets the UAV survey the SAR emulated area by following a lawnmower pattern.
    Offboard mode runs the POMDP motion planner by populating an initial victim position belief across the entire flying area.
    Hybrid mode extends the functionality of mission mode by running the POMDP motion planner to inspect the area delimited by the camera's FOV.}
  \label{fig:flight-modes}
\end{figure*}

\subsection{Flight Modes}
The proposed UAV system is evaluated by collecting data using three flight modes: \textit{mission}, \textit{offboard}, and \textit{hybrid}.
The survey extent for these tests is delimited by a 6 m $\times$ 60 m rectangular area, drafted in QGroundControl.
Specific details of the survey pattern can be found in Fig.~\ref{fig:flight_plan_rgb} and Tab.~\ref{tab:survey_params_rgb} from the Appendix.
A diagram illustrating the functionality of tested flight modes is shown in Fig.~\ref{fig:flight-modes}.

\subsubsection{Mission Mode}
When \textit{mission} mode is activated, the UAV automatically follows a list of position and velocity waypoints which define the survey plan previously drafted in QGroundControl and uploaded to the autopilot before starting the flight operation.
This flight mode is traditionally supported in many autopilots, and its out-of-the-box implementation serves as the planner baseline of this research.
While mission mode is operated in the UAV, the object detector is running in parallel to record any positive detections while the UAV is navigating in the environment and completing the survey.

\subsubsection{Offboard mode}
\textit{Offboard} mode offers autonomous navigation without a predefined survey plan of the environment.
This flight mode internally executes the POMDP-based motion planner described in \Cref{sec:planner} by declaring as flight parameters the initial position waypoint where the UAV should begin the survey, and the global coordinates of the survey extents.
The list of parameters can be found in Tab.~\ref{tab:initial_conds} from the Appendix.

\subsubsection{Hybrid Mode}
This paper proposes the fusion of the provided capabilities between mission and offboard modes, in a flight mode denominated \textit{hybrid}.
The aim of this flight mode is to take advantage of the initial awareness and survey coverage coming from mission mode in outdoor environments with GNSS signal coverage, and the autonomous navigation capabilities of offboard mode.
Instead of running the POMDP-based motion planner covering the entire extent of the surveyed area, in hybrid mode the survey extent is only limited by the extent of the camera's FOV.
Once a first detection is received from the vision module, this flight mode triggers offboard mode, boots the motion planner and passes action commands to the autopilot until the POMDP solver reaches a terminal state (\ie the UAV discards or confirms a victim).
Afterwards, the UAV resumes its survey by triggering back mission mode.
The process repeats itself with new detection outputs until the UAV completes the survey in mission mode.

\subsection{POMDP solver}
The navigation problem modelled as a POMDP is solved in real time through the use of the TAPIR toolkit~\cite{Klimenko2014}.
TAPIR is coded using the C++ programming language and encapsulates the Augmented Belief Trees (ABT) solver \cite{Kurniawati2016} to calculate and update the motion policy online.
ABT reduces computational demands by reusing past computed policies and updating the optimal policy if changes to the POMDP model are detected.
Furthermore, formulated problems with ABT allow declaring continuous values for actions, states, and observations.

Once the motion server calls the motion planner after a first victim detection is received by the object detector, TAPIR is booted by calculating an offline policy for four seconds.
Afterwards, the observation server retrieves an observation, updates the motion policy and takes the action that returns the highest expected reward.
An idle period of 3.4 seconds is applied for the UAV to reach the desired position coordinate, and then, the process repeats itself by requesting a new observation from the observation server.
The loop is broken once the detection confidence $\zeta$ exceeds a threshold.
Specific parameters from the TAPIR toolkit and ABT solver are shown in \cref{tab:initial_conds} from the Appendix.

\section{Results and Discussion}\label{sec:results}
The proposed UAV framework is evaluated through the performance indicators listed as follows:
(1) Successful detections per flight mode;
(2) Spatial distribution of recorded GNSS coordinates via heatmaps;
(3) Elapsed time taken by the UAV to locate the victim per location;
and
(4) Scalability test using thermal imagery.
Real flight demonstrations of the UAV framework can be found at \url{https://youtu.be/U_9LbNXUwV0}.

Accuracy metrics of victim detections were recorded using three variables: True Positives (TP), False Positives (FP), and False Negatives (FN).
TP is defined here as the relative number of flight runs where the victim was successfully detected at the true location.
FP is the relative number of flights which recorded victim locations in other areas than the true position of the victim.
FN is the relative number of flights that did not detect the victim at their real location.
In this context, a given flight test could report false positive detections and still detect the victim at the real location.
A summary table of collected metrics is depicted in Tab.~\ref{tab:metrics_rgb}.

\begin{table}[!tb]
  \caption{Accuracy metrics of the system to locate a victim at two locations (L1~and~L2).
    Here, TP are true positives, FP are false positives, and FN are false negatives.}
  \label{tab:metrics_rgb}
  \centering
  \small
  \begin{tabular}{|c|c|r|r|r|}
    \hline
    \bfseries Flight Mode & \bfseries Runs & \bfseries TP (\%) & \bfseries FP (\%) & \bfseries FN (\%) \\
    \hline\hline
    Mission (L1)          & \(5 \)         & \(100.0 \)        & \(20.0 \)         & \(0.0 \)          \\
    Offboard (L1)         & \(7 \)         & \(57.1 \)         & \(0.0\)           & \(42.9 \)         \\
    Hybrid (L1)           & \(5 \)         & \(80.0 \)         & \(0.0 \)          & \(20.0 \)         \\
    Mission (L2)          & \(5 \)         & \(60.0 \)         & \(0.0 \)          & \(40.0 \)         \\
    Offboard (L2)         & \(5 \)         & \(80.0\)          & \(0.0 \)          & \(20.0 \)         \\
    Hybrid (L2)           & \(7 \)         & \(71.4 \)         & \(28.6 \)         & \(28.6\)          \\
    \hline
  \end{tabular}
\end{table}

\begin{figure}[!tb]
  \begin{minipage}[c]{\columnwidth}
    \centering
    \includegraphics[width=0.89\columnwidth]{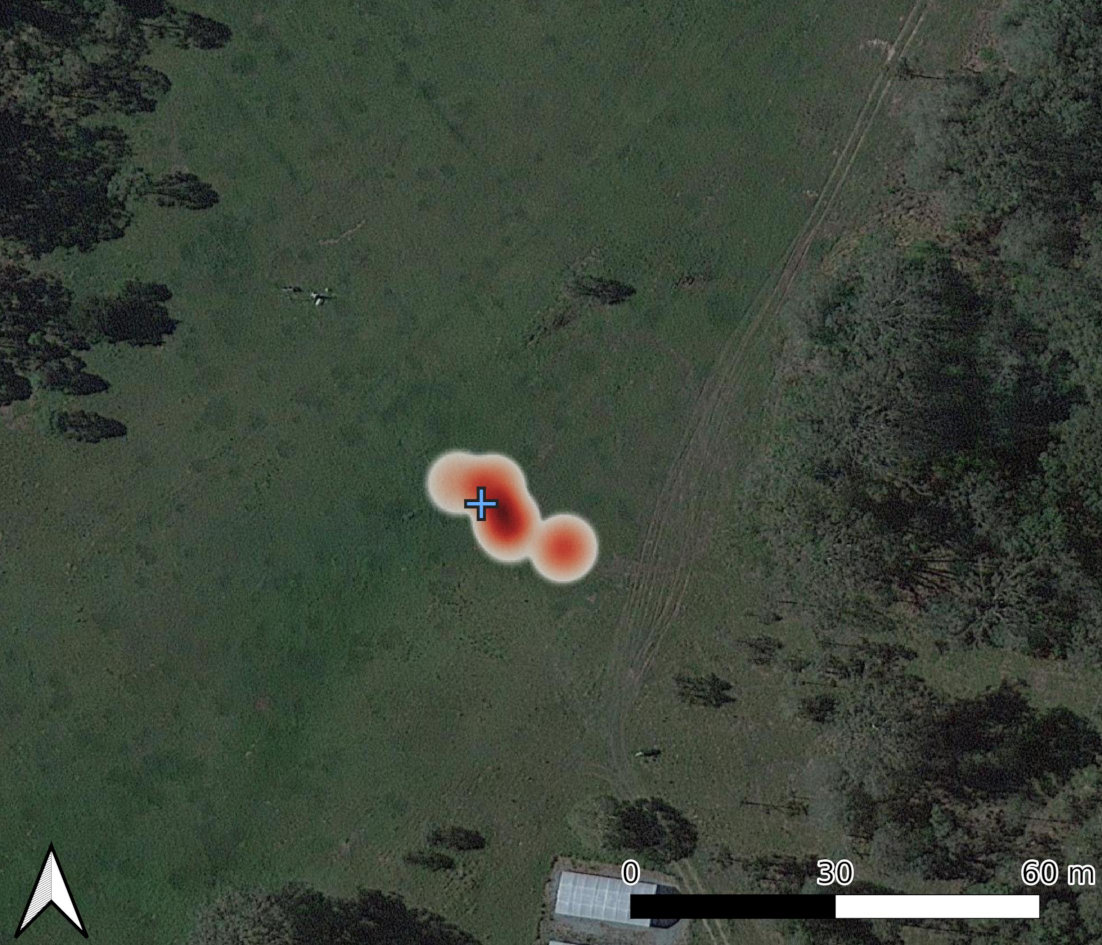}
    \\
    (\textbf{a})
  \end{minipage}
  \\
  \begin{minipage}[c]{\columnwidth}
    \centering
    \includegraphics[width=0.89\columnwidth]{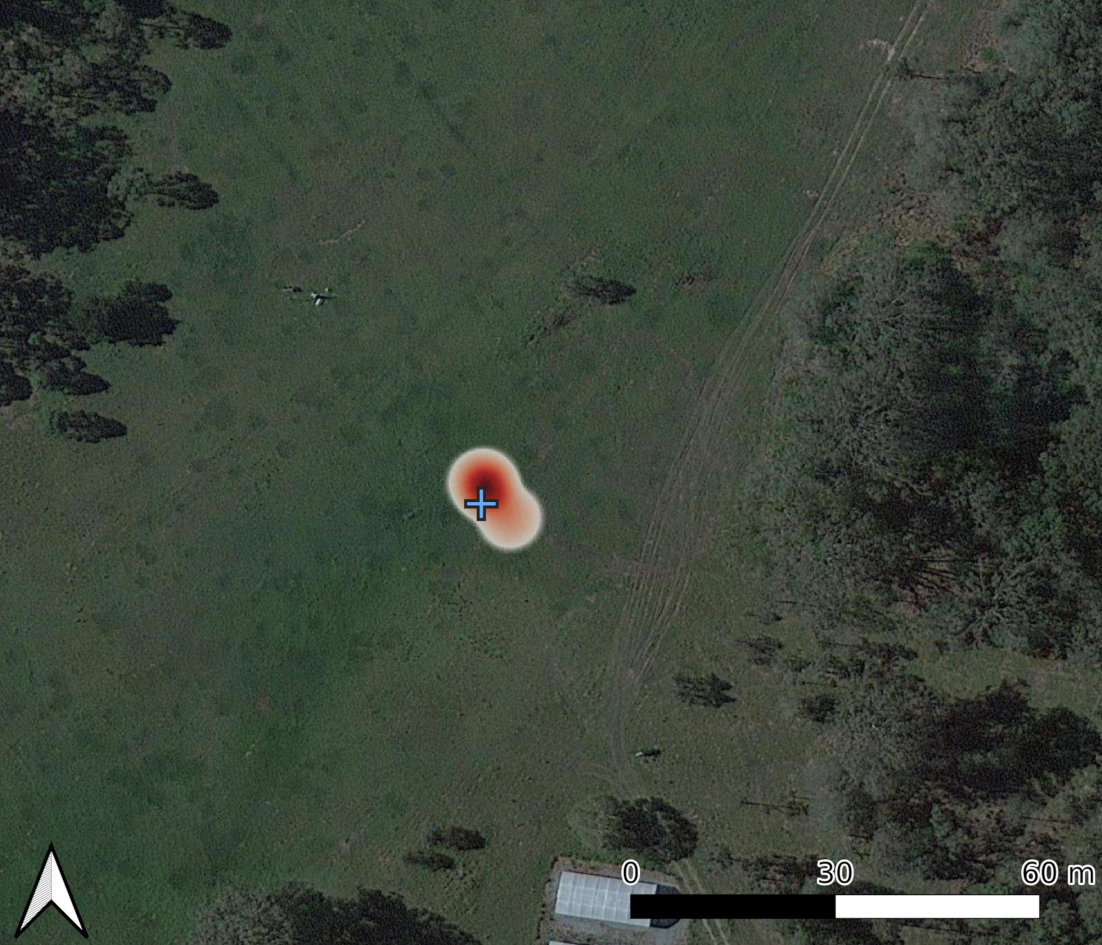}
    \\
    (\textbf{b})
  \end{minipage}
  \\
  \begin{minipage}[c]{\columnwidth}
    \centering
    \includegraphics[width=0.89\columnwidth]{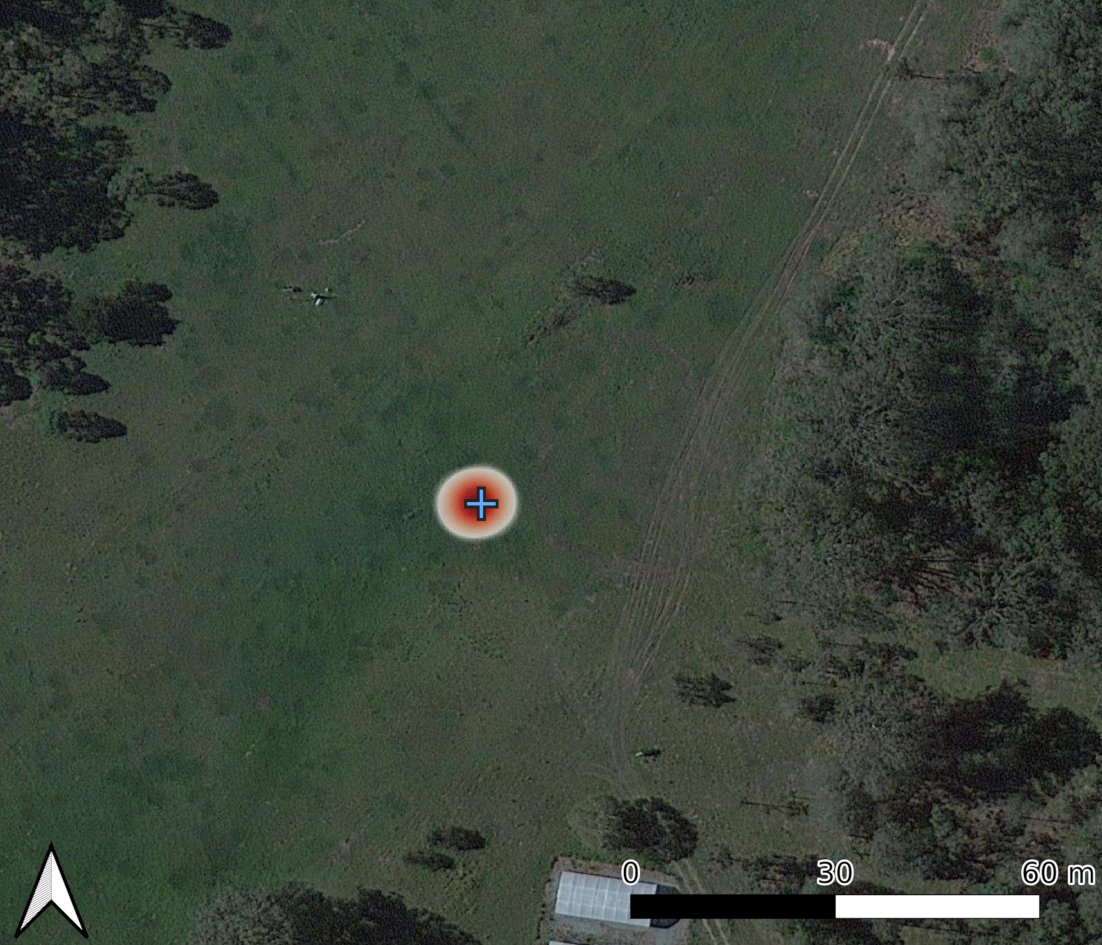}
    \\
    (\textbf{c})
  \end{minipage}
  \caption{Heatmaps of recorded GNSS coordinates in a trivial victim location (L1) using (a) mission, (b) offboard, and (c) hybrid flight modes.}
  \label{fig:rgb_heatmap_loc_1}
\end{figure}

Accuracy metrics of the proposed framework provided contrasting results at the tested victim locations.
On flight tests with the victim placed in a trivial location (\ie, L1), the UAV achieved 100\% of positive victim detections in mission mode, and 20\% of those recorded GNSS coordinates of false victim locations.
For tests in offboard and hybrid flight modes, the true positive rates decreased in comparison with mission mode.
However, both setups achieved flight tests without any false positive readings.
An illustration of the spatial distribution of recorded GNSS coordinates per flight mode for L1 is shown in Fig.~\ref{fig:rgb_heatmap_loc_1}.

\begin{figure}[!tb]
  \centering
  \includegraphics[width=\columnwidth]{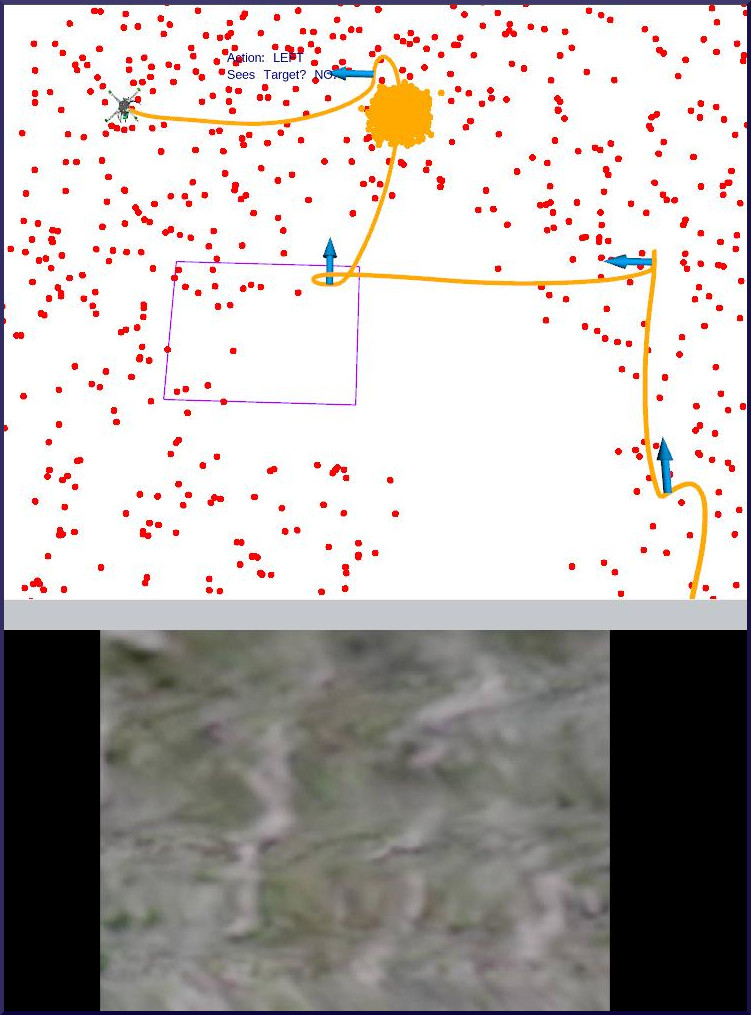}
  \vspace{-3ex}
  \caption{Strong winds distorting RGB streaming in offboard mode.
    The top image illustrates the traversed path of the UAV and the probability distribution of the UAV and victim locations (orange and red points respectively).
    The bottom image shows the latest (distorted) streamed frame from the RGB camera.}
  \label{fig:offboard_windy}
\end{figure}

Flight tests with the mannequin located in a complex location (\ie, L2) showed an overall improvement in TP rates for offboard and hybrid modes compared to the baseline planner (mission mode).
For mission and offboard modes, there were no flights which reported FP detections even though the flight setup, payload, and object detector remained the same while testing the framework with the mannequin located in L1.
Nevertheless, 28.6\% of the flights in hybrid mode reported false positive victim locations.
The rate of false positives for all flight tests were caused by limitations from the object detector by recording other objects as humans (\ie, from the car placed in the flight area).
Conversely, most of the false negative records during the flight tests occurred from excessive vibration in the UAV frame caused by strong winds, as shown in Fig.~\ref{fig:offboard_windy}.

\begin{figure}[!tb]
  \begin{minipage}[c]{\columnwidth}
    \centering
    \includegraphics[width=0.89\columnwidth]{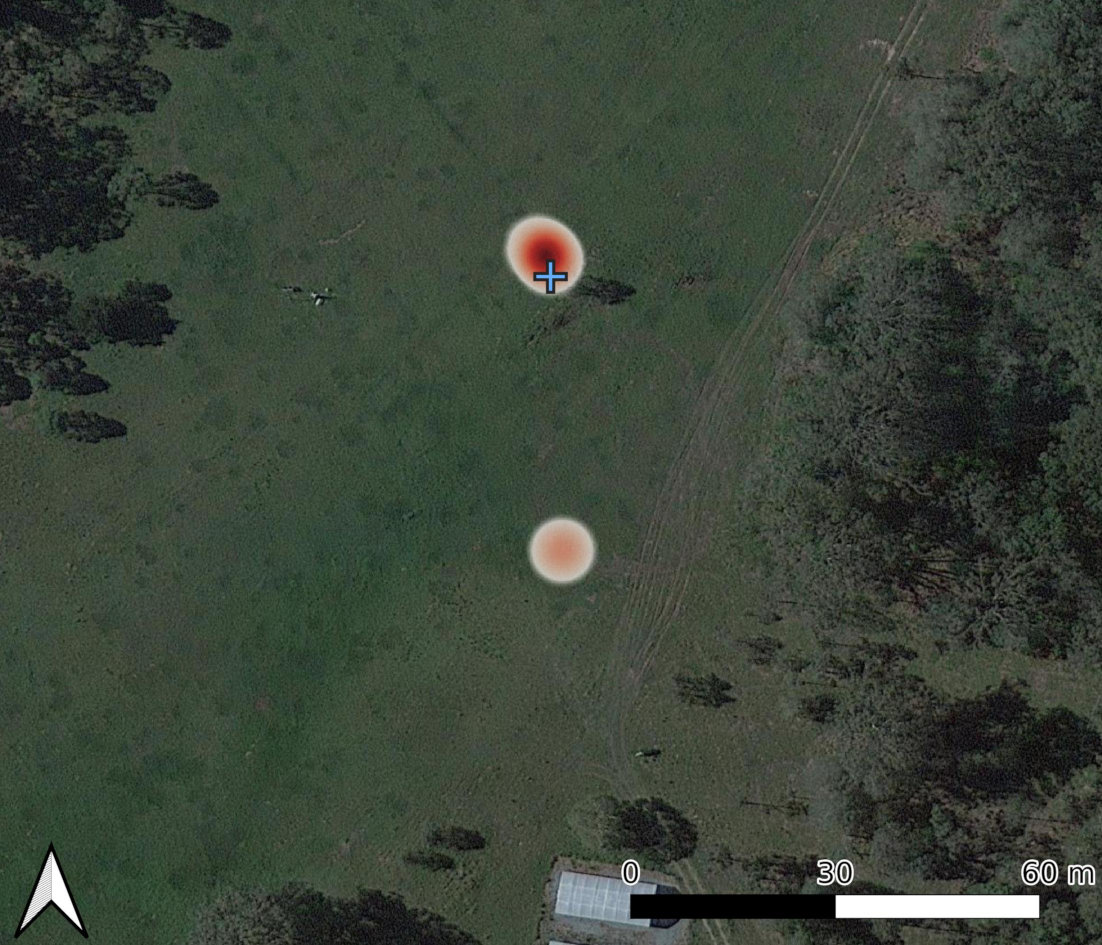}
    \\
    (\textbf{a})
  \end{minipage}
  \\
  \begin{minipage}[c]{\columnwidth}
    \centering
    \includegraphics[width=0.89\columnwidth]{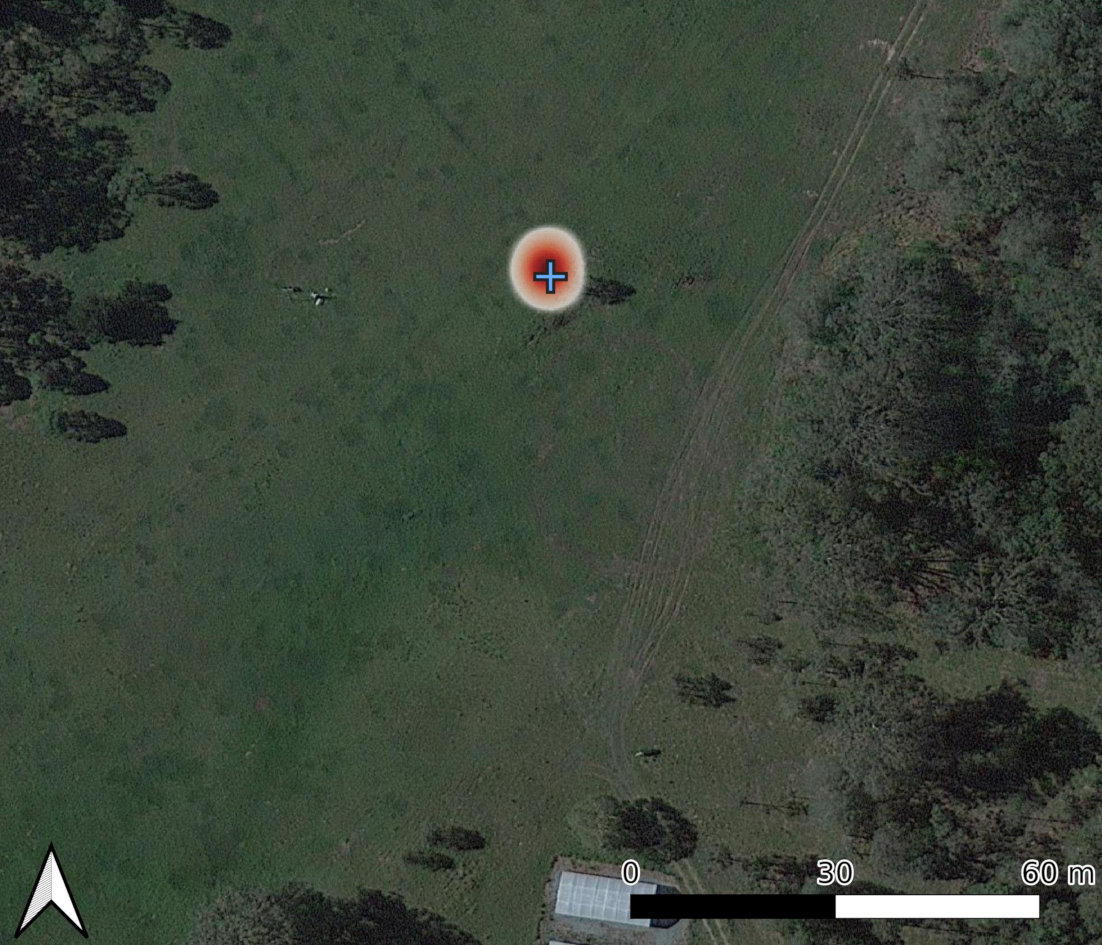}
    \\
    (\textbf{b})
  \end{minipage}
  \\
  \begin{minipage}[c]{\columnwidth}
    \centering
    \includegraphics[width=0.89\columnwidth]{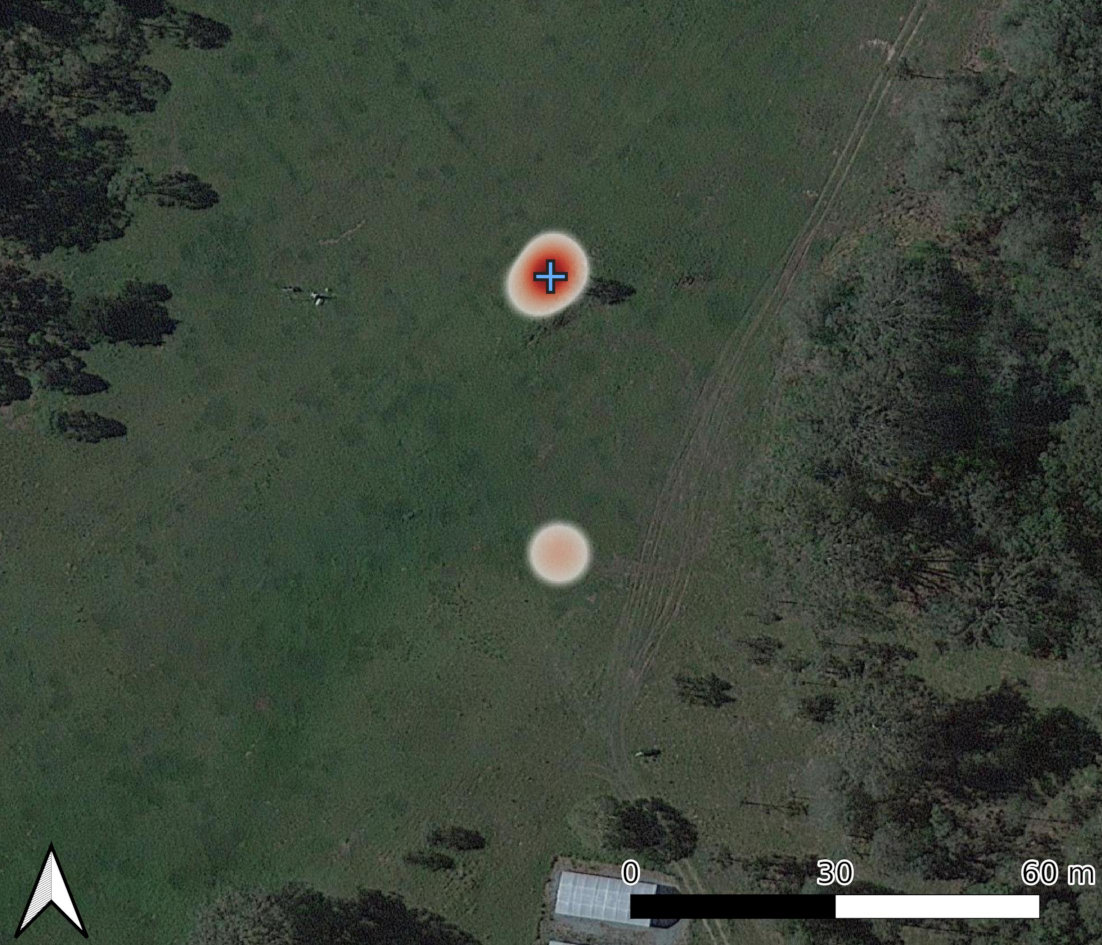}
    \\
    (\textbf{c})
  \end{minipage}
  \caption{Heatmaps of recorded GNSS coordinates in a complex victim location (L2) using (a) mission, (b) offboard, and (c) hybrid flight modes.}
  \label{fig:rgb_heatmap_loc_2}
\end{figure}

A visual analysis of the spatial distribution of recorded GNSS coordinates during the flight tests is performed using heatmaps.
The heatmaps, illustrated in Figs.~\ref{fig:rgb_heatmap_loc_1} and~\ref{fig:rgb_heatmap_loc_2}, indicate a reduction in victim location uncertainty after operating the proposed UAV framework using offboard and hybrid modes.

\begin{figure}[!tb]
  \begin{minipage}[c]{\columnwidth}
    \centering
    \includegraphics[width=0.95\columnwidth]{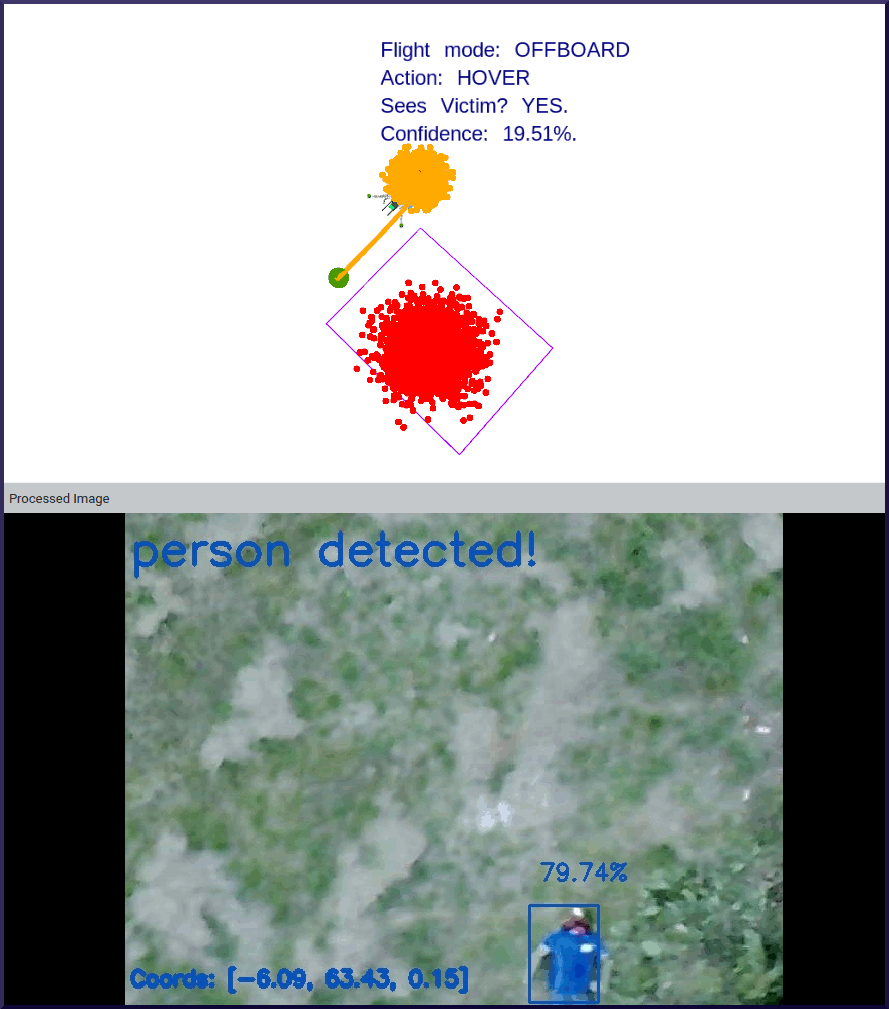}
    \\
    (\textbf{a})\\
    ~\\
    ~\\
  \end{minipage}
  \\
  \begin{minipage}[c]{\columnwidth}
    \centering
    \includegraphics[width=0.95\columnwidth]{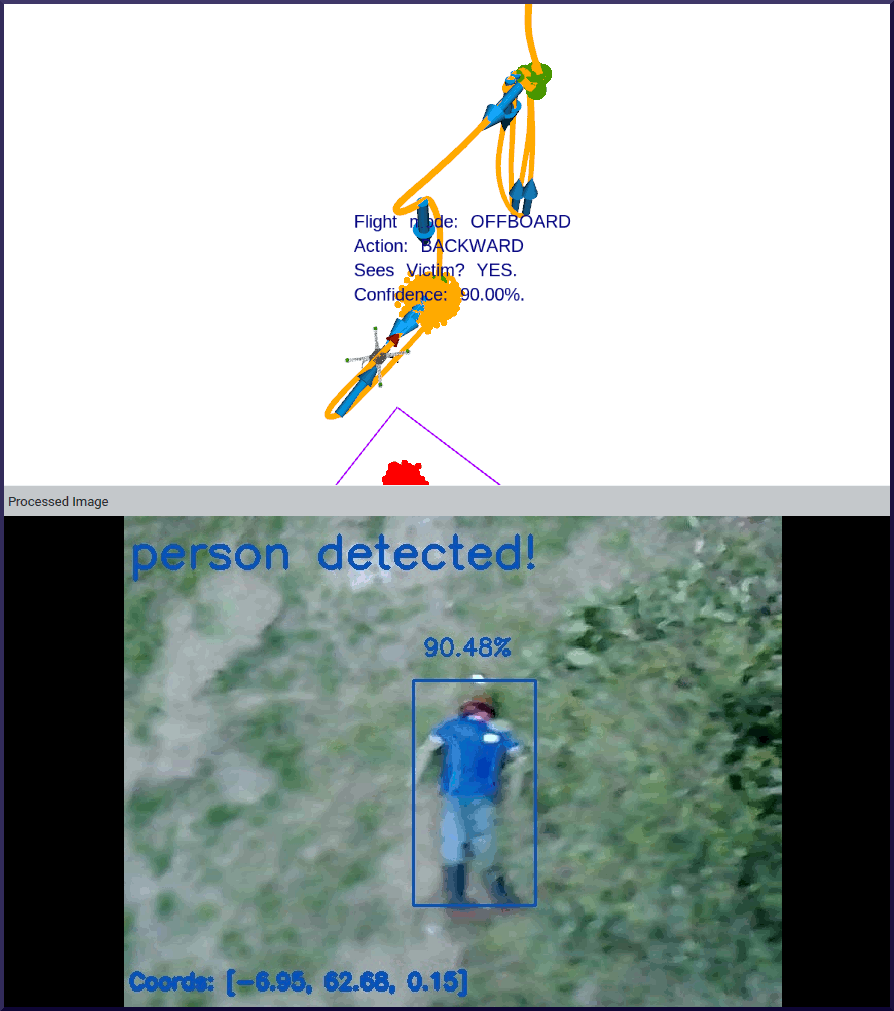}
    \\
    (\textbf{b})\\
  \end{minipage}
  \centering
  \caption{Reduction of object detection uncertainty from RGB camera after executing the motion policy.
    (a) Initial detection of a potential victim with a confidence value of 19.51\%.
    (b) Increased confidence value (90.0\%) at subsequent detections after the UAV executes actions from the computed motion policy and gets closer to the victim.}
  \label{fig:rgb_confidence}
\end{figure}

The presented POMDP-based motion planner from~\cite{Sandino2020a}---scaled in this work for UAV navigation in outdoor environments---contributed on the reduction of object detection uncertainty in flight tests using offboard and hybrid modes.
An example of how the UAV inspects an area to confirm whether a victim is truly located after receiving an initial detection is shown in Fig.~\ref{fig:rgb_confidence}.
In the first time steps (Fig.~\ref{fig:rgb_confidence}a), a low confidence value of 19.51\% is retrieved because of the few number of pixels representing the mannequin while surveying at 16 m, and partial occlusion from the nearby tree.
After taking actions commands from the POMDP policy computed in the planner module, the UAV is positioned closer to the mannequin (\ie, 10 m) and with a better viewpoint of the scene, retrieving a confidence value of 90.0\% (Fig.~\ref{fig:rgb_confidence}b).
The traversed path by the UAV also suggests the capability of the UAV to adapt (or update) its motion policy while it interacts with the environment and receives new observations.
Adjustments in the motion policy also occur from uncertainty sources, such as unexpected strong wind currents, oscillating GNSS signal errors, illumination changes, and false detections by the CNN model.

\begin{figure}[!tb]
  \centering
  \includegraphics[width=\columnwidth]{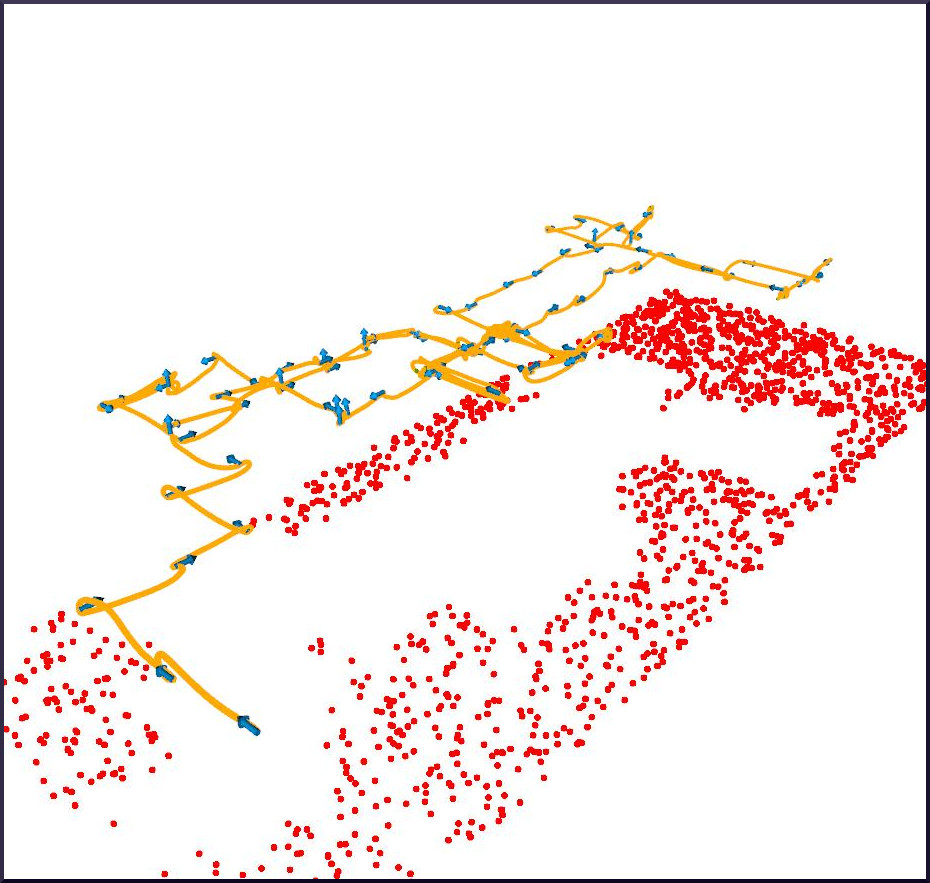}
  \vspace{-3ex}
  \caption{Example traversed path in offboard mode while no victims are found.
    The UAV moves towards the centre of the surveyed area before exploring its corners.}
  \label{fig:offboard_missed}
\end{figure}

This work also studied the speed of the proposed framework to find victims using mission, offboard, and hybrid flight modes.
As presented in Tab.~\ref{tab:duration_rgb}, two primary insights are observed in elapsed times between victim locations and flight modes.
The first observed trend defines higher values of standard deviation (especially in offboard mode) in flight tests with the victim at location L1.
These values were caused by slight inconsistencies in the traversed path from the motion planner, as seen in Fig.~\ref{fig:offboard_missed}.
From the surveyed area, Location 1 is closer to the survey limits, whereas Location 2 is placed closed the upper centre.
The motion planner lets the UAV move towards the centre of the surveyed area before exploring its corners.
Despite test runs in offboard mode providing higher overall accuracy values than tests in hybrid mode, surveyed patterns could become suboptimal if a victim is located close to the boundaries of the delimited flying area.

\begin{table}[!tb]
  \caption{Elapsed time by the UAV to locate a victim at two locations (L1 and L2) per flight mode.
    Here, SD stands for Standard Deviation and SE stands for Standard Error.}
  \label{tab:duration_rgb}
  \centering
  \small
  \begin{tabular}{|c|c|c|c|}
    \hline
    \bfseries Flight Mode & \bfseries Mean (s) & \bfseries SD (s) & \bfseries SE (s) \\
    \hline\hline
    Mission               & \(169.67\)         & --               & --               \\
    Offboard (L1)         & \(146.24 \)        & \(128.70\)       & \(64.35\)        \\
    Hybrid (L1)           & \(392.39 \)        & \(130.20\)       & \(65.10\)        \\
    Offboard (L2)         & \(148.30 \)        & \(25.58\)        & \(12.79\)        \\
    Hybrid (L2)           & \(289.26 \)        & \(76.84\)        & \(34.36\)        \\
    \hline
  \end{tabular}
  \vspace{1ex}
\end{table}

The second observed trend is that hybrid mode recorded longer times for detecting and confirming the victim, regardless of its location in the surveyed area.
This impact highly depends on the recall properties of the vision-based object detector.
The number of false positive outputs while the UAV explores the environment defines the number of inspections, which will increase the flight time until the survey is complete.
Using other object detectors tuned from airborne UAV datasets is expected to reduce the survey duration in hybrid mode, as these models should provide higher recall values than the MobileNet SSD detector implemented in this paper.

Limitations in the implemented object detector to output positive detections have conditioned the maximum altitude for the surveys, and consequently, impacted the overall speed of the system to complete the mission.
Other restrictions are defined by the gimbal configuration, which scopes the footprint of the camera's FOV, and the image resolution of streamed frames.
Therefore, reduced times to accomplish the victim finding mission can be accomplished by improvements in the object detector, higher image resolution and oblique camera angles that provide a wider viewpoint of the scene.
Nonetheless, such improvements might involve an increment in computational power and further research should evaluate any impacts from better detectors and processing of high-resolution frames in embedded systems such as the UP\textsuperscript{2}.

The scalability test of the presented framework was achieved with preliminary flight tests using thermal imagery.
The modular components replaced for this trial consisted of the UAV payload, the vision-based detector, and the flight plan generated by QGroundControl to match the desired overlap values previously tested with the RGB camera.

The object detector is a custom implementation of a TinyYoloV2 model architecture from Microsoft Azure Custom Vision services.
A total of 5175 labelled images were used in the training process.
The thermal dataset contains images of adults with a rich set of posing configurations: adults lying on the ground and waving their arms; adults standing up waving their arms; altitude and camera viewpoint variations from the UAV, ranging from 5 to 40 m, and gimbal angles of 45{\textdegree} and 0{\textdegree} from Nadir.

\begin{figure}[!tb]
  \begin{minipage}[c]{\columnwidth}
    \centering
    \includegraphics[width=\columnwidth]{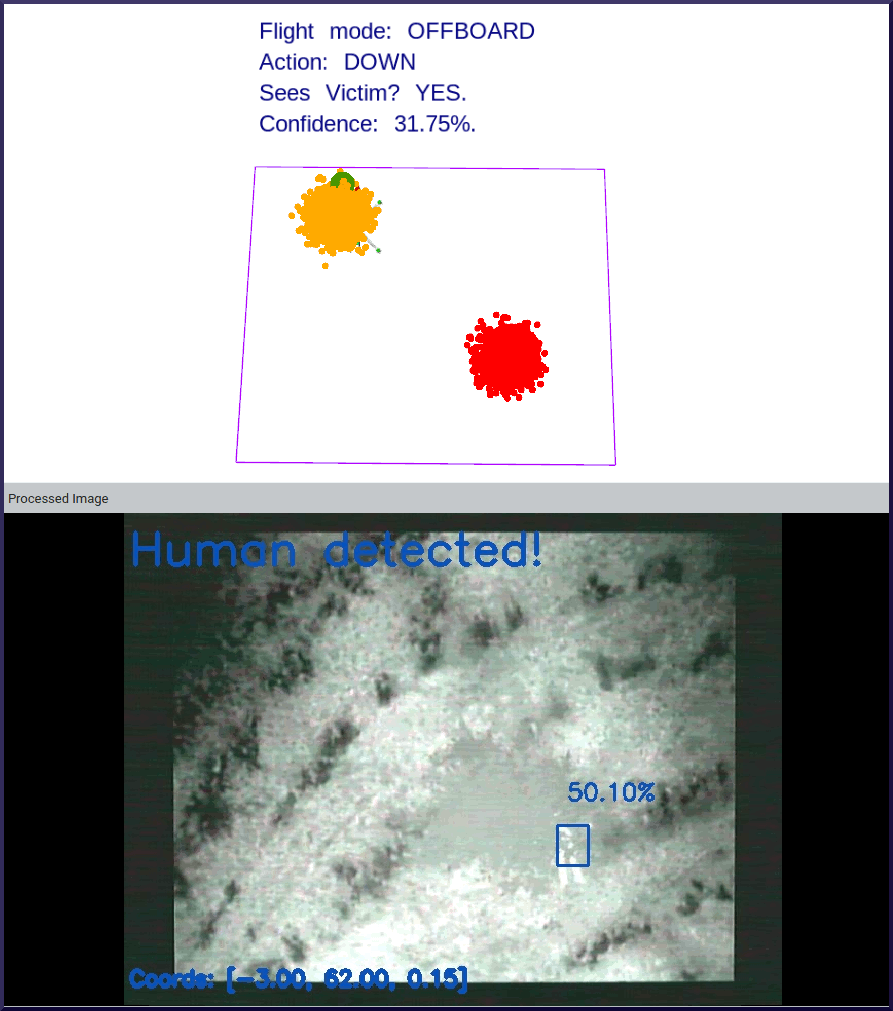}
    \\
    (\textbf{a})\\
  \end{minipage}
  \\
  \begin{minipage}[c]{\columnwidth}
    \centering
    \includegraphics[width=\columnwidth]{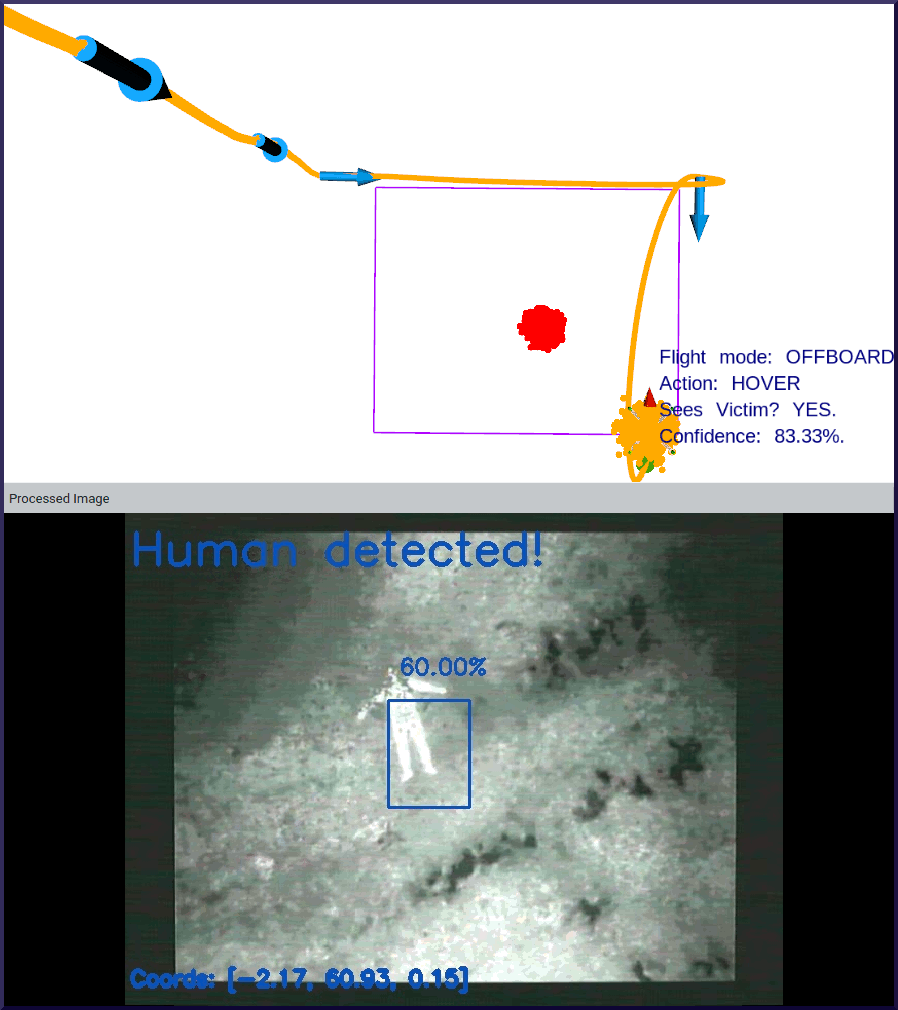}
    \\
    (\textbf{b})
  \end{minipage}
  \caption{Reduction of object detection uncertainty from thermal camera after executing the motion policy.
    (a) Initial detection of a potential victim with a confidence value of 31.75\%.
    (b) Increased confidence value (83.33\%) at subsequent detections after the UAV gets closer to the victim.}
  \label{fig:thermal_confidence}
\end{figure}

Thermal preview tests were conducted between 5:30 a.m. and 7:30 a.m. from 25 Aug 2021 to 23 Sep 2021, with ambient temperatures between 7{\textdegree}C and 14{\textdegree}C.
An overview of the system navigating in hybrid mode to increase values of object detection confidence from streamed thermal frames can be found in Fig.~\ref{fig:thermal_confidence}.
In this instance, the UAV starts inspecting the area covered by the FOV of the thermal camera (Fig.~\ref{fig:thermal_confidence}a).
In subsequent time steps, the aircraft gets closer to have a better visual of the scene and confirm the presence of the victim by retrieving higher detection confidence values (Fig.~\ref{fig:thermal_confidence}b).
A~demonstrative preview video of flight operations with thermal imagery can be found at \url{https://youtu.be/yIPNBwNYtAo}.

The proposed UAV framework constitutes a novel approach to autonomous navigation for exploration and target finding under uncertainty.
Real-world environments are full of uncertainties such as illumination conditions, strong wind currents, collision from static and dynamic obstacles, occlusion, and limitations in object detectors, which can negatively affect the success of the flight mission.
While many approaches handle object detection uncertainty by fine-tuning CNN models with labelled airborne datasets \cite{Mittal2020},
this research suggests augmenting the cognition power onboard UAVs from imperfect sensor and detection output observations.
Moreover, augmenting autonomy capabilities in small UAVs might open more approaches to automated surveying in outdoor environments by using a swarm of UAVs which will not require permanent supervision by pilots to photo-interpret streamed camera frames to identify and locate potential victims.

Several benefits can be offered through the use of the framework in time-critical applications to SAR squads.
A first assessment of the accessibility conditions of the surveyed environment, identification, localisation, quantification and conditions of victims and external hazards might be obtained rapidly thanks to real-time telemetry of processed camera frames.
After completing a flight, the list of GNSS coordinates (depicted in Figs.~\ref{fig:rgb_heatmap_loc_1} and~\ref{fig:rgb_heatmap_loc_2}) can be shared to SAR squads to coordinate better response strategies, as well as launching additional UAVs for critical deployment of medicines, food or water.

\section{Conclusions}\label{sec:conclusions}
This paper discussed a modular UAV framework for autonomous onboard navigation in outdoor environments under uncertainty.
The system showed how levels of object detection uncertainty were substantially reduced by calculating a motion policy using an online POMDP solver and interacting with the environment to obtain better visual representations of potential detected targets.
CNN-based computer vision inference and motion planning can be executed in resource-constrained hardware onboard small UAVs.
The framework design was validated with real flight tests with a simulated SAR mission, which consisted in finding an adult mannequin in an open area and close to a tree.
Collected performance indicators from three flight modes suggest that the system reduces levels of object detection uncertainty in outdoor environments whether information about the surveyed environment is available.
This framework was also extended by adapting the payload with a thermal camera and converting a customised people detector from thermal imagery in the vision module.

The presented framework extends the contributions in \cite{Sandino2020,Sandino2021} by:
(1) extending their tested UAV framework in GNSS-denied environments for outdoor missions with GNSS signal coverage with a novel flight mode (\ie, hybrid mode),
(2) further validating preliminary results of the presented framework using real flight tests,
and
(3) demonstrating scalability opportunities of the modular framework design by adapting a thermal camera and custom object detector to locate victims using their heat signatures.

Future work should evaluate the performance of the UAV system using tailored object detectors from network architectures different than MobileNet SSD and high-resolution streamed frames.
Further assessments with multiple victims and more complex environment configurations (\eg~slope terrain, type and number of obstacles) are encouraged.
A comparison study of various online POMDP solvers, or other motion planners, could improve understanding the limits of using the ABT solver and POMDP solvers for motion planning under uncertainty.


\appendix{}              

\begin{figure}[!t]
  \centering
  \includegraphics[width=0.9\columnwidth]{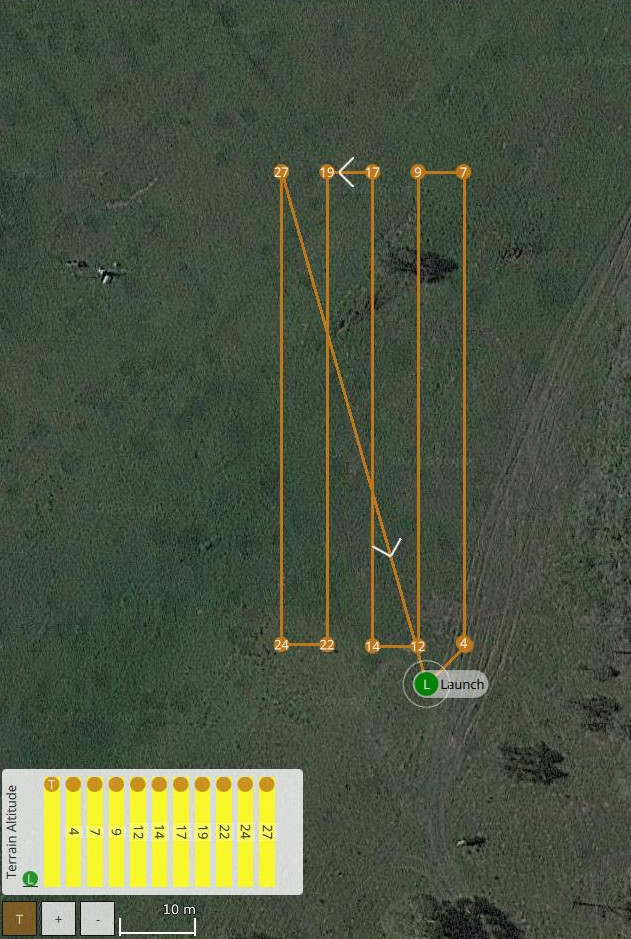}
  \caption{Flight plan following a lawnmower pattern using the RGB camera properties from Tab.~\ref{tab:survey_params_rgb}.}
  \label{fig:flight_plan_rgb}
\end{figure}

The survey design and flight plan parameters for conducted tests in \textit{mission} and \textit{hybrid} flight modes are shown in Fig.~\ref{fig:flight_plan_rgb} and Tab.~\ref{tab:survey_params_rgb}, respectively.

\begin{table}[!t]
  \caption{Flight plan parameters based on RGB camera properties.}
  \label{tab:survey_params_rgb}
  \small
  \centering
  \begin{tabular}{|l|l|}
    \hline
    \bfseries Property    & \bfseries Value                                  \\ \hline\hline
    UAV altitude          & 16 m                                             \\
    UAV velocity          & 2 m/s                                            \\
    Lens width            & 2.06 mm                                          \\
    Lens height           & 1.52 mm                                          \\
    Camera focal length   & 4.7 mm                                           \\
    Image resolution      & 640 by 480 px                                    \\
    Overlap               & 30\%                                             \\
    Bottom right waypoint & -27.3892746\textdegree, 152.8730164\textdegree \\
    Top right waypoint    & -27.3887138\textdegree, 152.8730164\textdegree \\
    Top left waypoint     & -27.3887138\textdegree, 152.8727722\textdegree \\
    Bottom left waypoint  & -27.3892765\textdegree, 152.8727722\textdegree \\
    \hline
  \end{tabular}
\end{table}

The set of hyper-parameters used in TAPIR and initial conditions to operate the UAV using offboard and hybrid modes are shown in Tab.~\ref{tab:initial_conds}.

\begin{table}[!t]
  \caption{Set of hyper-parameters used in TAPIR and initial conditions to operate the UAV in offboard and hybrid flight modes.}
  \label{tab:initial_conds}
  \centering
  \small
  \begin{tabular}{cll}
    \hline
    \bfseries Variable     & \bfseries Description        & \bfseries Value                                          \\ \hline\hline
    \(z_{\text{max}} \)    & Maximum UAV altitude         & 16 m                                                     \\
    \(z_{\text{min}} \)    & Minimum UAV altitude         & 5.25 m                                                   \\
    \(p_{u0} \)            & Initial UAV position         & {\tiny(-27.3897972\textdegree, 152.8732300\textdegree, 20m)} \\
    \(\varphi_{u} \)       & UAV Heading                  & 0\textdegree                                            \\
    \(\delta_z\)           & UAV climb step               & 2 m                                                      \\
    \(\lambda\)            & Frame overlap                & 40\%                                                     \\
    \(\alpha\)             & Camera pitch angle           & 0\textdegree                                            \\
    \(\zeta_{\text{min}}\) & Minimum detection confidence & 10\%                                                     \\
    \(\zeta\)              & Confidence threshold         & 85\%                                                     \\
    \(\gamma\)             & Discount factor              & 0.95                                                     \\
    \(\Delta t\)           & Time step interval           & 4 s                                                      \\
  
    \(t_{\text{max}}\)     & Maximum flying time          & 10 min                                                   \\
    \hline
  \end{tabular}
\end{table}



\section*{Acknowledgements}
\begin{small}

The authors wish to express their gratitude to Sharlene Lee-Jendili for her implementation of the thermal-based people detector used in this project.
The authors acknowledge continued support from the Queensland University of Technology (QUT) through the Centre for Robotics.
Special thanks to the Samford Ecological Research Facility (SERF) team (Marcus Yates and Lorrelle Allen) for their continuous assistance and equipment provided during the flight tests.
The authors would also like to gratefully thank the QUT Research Engineering Facility (REF) team (Dr Dmitry Bratanov, Gavin Broadbent, Dean Gilligan) for their technical support that made possible conducting the flight tests.

This research was funded by the Commonwealth Scientific and Industrial Research Organisation (CSIRO) through the CSIRO Data61 PhD and Top Up Scholarships (Agreement 50061686);
the Australian Research Council (ARC) through the ARC Discovery Project 2018 ``Navigating under the forest canopy and in the urban jungle'' (grant number ARC DP180102250); and
the Queensland University of Technology (QUT) through the Higher Degree Research (HDR) Tuition Fee Sponsorship.
Special thanks to Hexagon through the Hexagon SmartNet RTK corrections service that enabled high accuracy surveying and positioning data using the EMLID Reach RTK receiver during the experimentation phase.

\end{small}

\small

\bibliographystyle{ieee_mod}
\bibliography{IEEEabr,references.bib}


\end{document}